\documentclass[10pt]{article} 
\usepackage[accepted]{tmlr}

\usepackage{amsmath,amsfonts,bm}

\def\eqref#1{equation~\ref{#1}}

\def\1{\bm{1}}

\DeclareMathAlphabet{\mathsfit}{\encodingdefault}{\sfdefault}{m}{sl}
\SetMathAlphabet{\mathsfit}{bold}{\encodingdefault}{\sfdefault}{bx}{n}

\usepackage[hidelinks]{hyperref}
\usepackage{url}
\usepackage{graphicx}
\usepackage{xspace}
\usepackage{tcolorbox}
\usepackage{lscape}
\usepackage{subcaption}
\usepackage{booktabs}
\usepackage{pifont}
\usepackage{multirow}
\usepackage{enumitem}
\usepackage{xcolor}
\usepackage{afterpage}
\newlength{\dpcircle}
\newlength{\rcircle}
\newlength{\dcircle}
\usepackage[normalem]{ulem}

\newcommand{\docircle}[4]{%
  \setlength{\dpcircle}{\dp\strutbox}%
  \setlength{\dcircle}{\dpcircle}%
  \addtolength{\dcircle}{\ht\strutbox}%
  \setlength{\rcircle}{0.5\dcircle}%
  \setlength{\unitlength}{1sp}%
  \begin{picture}(\number\dcircle,0)
    \color{#1}
    \put(\number\rcircle,\number\dpcircle){\circle*{\number\dcircle}}
    \color{#2}
    \put(\number\rcircle,\number\dpcircle){\circle{\number\dcircle}}
    \put(\number\rcircle,0){\makebox[0pt]{\textcolor{#3}{#4}}}
  \end{picture}%
}

\newcommand{\compiletask}{Error Detection\xspace}
\newcommand{\compiletasklower}{error detection\xspace}
\newcommand{\clonetask}{Clone Detection\xspace}
\newcommand{\clonetasklower}{clone detection\xspace}
\newcommand{\tagtask}{Solution Domain Classification\xspace}
\newcommand{\tagtasklower}{solution domain classification\xspace}
\newcommand{\refinetask}{Code Repair\xspace}
\newcommand{\refinetasklower}{code repair\xspace}
\newcommand{\cmark}{\ding{51}}%
\newcommand{\xmark}{\ding{55}}%
\newcommand{\LlamaInstr}{\textsc{Llama}~3.3~70B-Instruct\xspace}

\title{Cross-lingual Transfer in Programming Languages:\\An Extensive Empirical Study}

\author{\name Razan Baltaji \email baltaji@illinois.edu\\
      \addr Department of Electrical and Computer Engineering \\
      University of Illinois at Urbana-Champaign 
      \AND
      \name Saurabh Pujar\email saurabh.pujar@ibm.com \\
      \name Louis Mandel \email lmandel@us.ibm.com\\ 
      \name  Martin Hirzel \email hirzel@us.ibm.com \\ 
      \name Luca Buratti \email luca.buratti2@ibm.com \\
      \addr IBM Research
      \AND
    \name  Lav R. Varshney \email varshney@illinois.edu\\
        \addr Department of Electrical and Computer Engineering \\
      University of Illinois at Urbana-Champaign}

\def\month{06}  
\def\year{2025} 
\def\openreview{\url{https://openreview.net/forum?id=1PRBHKgQVM}}

\begin{document}

\maketitle

\begin{abstract}
Large language models (LLMs) have achieved state-of-the-art performance in various software engineering tasks, including error detection, clone detection, and code translation, primarily leveraging high-resource programming languages like Python and Java. However, many critical languages, such as COBOL, as well as emerging languages, such as Rust and Swift, remain low-resource due to limited openly available code. This scarcity hampers the training and effectiveness of LLMs for these languages, increasing software maintenance costs and stifling innovation. Addressing this gap, we investigate the potential of transfer learning to enhance LLM performance on low-resource programming languages by leveraging data from high-resource counterparts. Our extensive empirical study evaluates transferability across 10 to 41 programming languages and five key tasks: code generation, clone detection, code repair, solution domain classification, and error detection. Additionally, we develop a performance prediction model to guess the best source languages for a given target and task, and analyze the features that influence transfer performance. We further replicate a representative subset of experiments with a larger model to test the generalizability of our conclusions to contemporary large‑scale LLMs. Our findings demonstrate that cross-lingual transfer significantly outperforms zero-shot learning, with effectiveness varying based on both source and target languages. Languages such as Java and Go emerge as the best targets, while Kotlin and JavaScript are excellent sources. Furthermore, our model reliably predicts successful transfer sources by considering linguistic and dataset-specific features, offering practical guidance for data acquisition and model training. This work contributes to the development of LLM-driven tools for low-resource programming languages and provides insights into the characteristics that facilitate  transfer  across  language pairs.
\end{abstract}

\section{Introduction}
\label{sec:intro}

Large language models {(}LLMs{)} leverage the {fact that software code is often readable and repetitive (much like natural language}~\citep{hindle2016naturalness}) to achieve state-of-the-art performance on many software engineering tasks such as error detection, clone detection, and code translation~\citep{lu2021codexglue}. 
However, LLMs have thus far been trained and evaluated mainly on
\emph{high-resource} programming languages such as Python and Java with
vast amounts of openly available code.
Yet there are many \emph{low-resource} programming
languages that lack enough openly available code
to train LLMs.

COBOL, which supports many critical financial applications and may have over 775 billion lines of code overall~\citep{cobolmarket}, is low-resource due to lack of open availability. Notwithstanding their popularity~\citep{zhou2022improving,courseraswift}, new languages such as Rust and Swift start out low-resource, by definition. Other languages such as R are low-resource because most code in that language has a license that inhibits use for LLM training~\citep{li_et_al_2023}. Besides being underrepresented in training of LLMs, organizations often spend  significant resources on training software developers to work with unfamiliar low-resource languages. This adds to the cost of software maintenance and may also stifle innovation \citep{EdwardsK2021}. There is a need for state-of-the-art AI tools to enhance developer productivity for low-resource languages.

Recent works have shown the possibility of transfer learning: leveraging data from one programming language to compensate for the lack of data in another~\citep{ahmed2022multilingual,yuan_et_al_2022,pian_et_al_2023,cassano_et_al_2024-oopsla}.
Another recent work has developed several similarity metrics to decide which high-resource programming language dataset can be used to augment fine-tuning data for a target language~\citep{chen2022transferability}. 
Prior work, however, has used at most 6~programming languages, none of which are truly low-resource.
Further, the notion of similarity among  programming languages is underexamined: more research is needed to make reliable claims.

To overcome these shortcomings, we examine transferability from
source to target language combinations for 5~tasks and for many
more languages than prior work (10--41 depending on the task).
We start by using training data for a task in a source language to
fine-tune a small LLM~(for 4 tasks) or to perform in-context
learning using a larger LLM~(for the 5th task).
Then, we feed the respective LLM disjoint test data for the same
task on a target language.
Using this methodology, we evaluate the results across all
source languages, target languages, and tasks.
The source languages are, by definition, high-resource languages
since we have enough data to train a model on them.
Target languages include both high-resource and low-resource ones.
The finetuning experiments focus on the four tasks
\clonetasklower, \refinetasklower, \tagtasklower, and \compiletasklower.
The in-context learning experiments use a fifth task, code generation.
\begin{figure*}[!t]
  \centerline{\includegraphics[width=\textwidth]{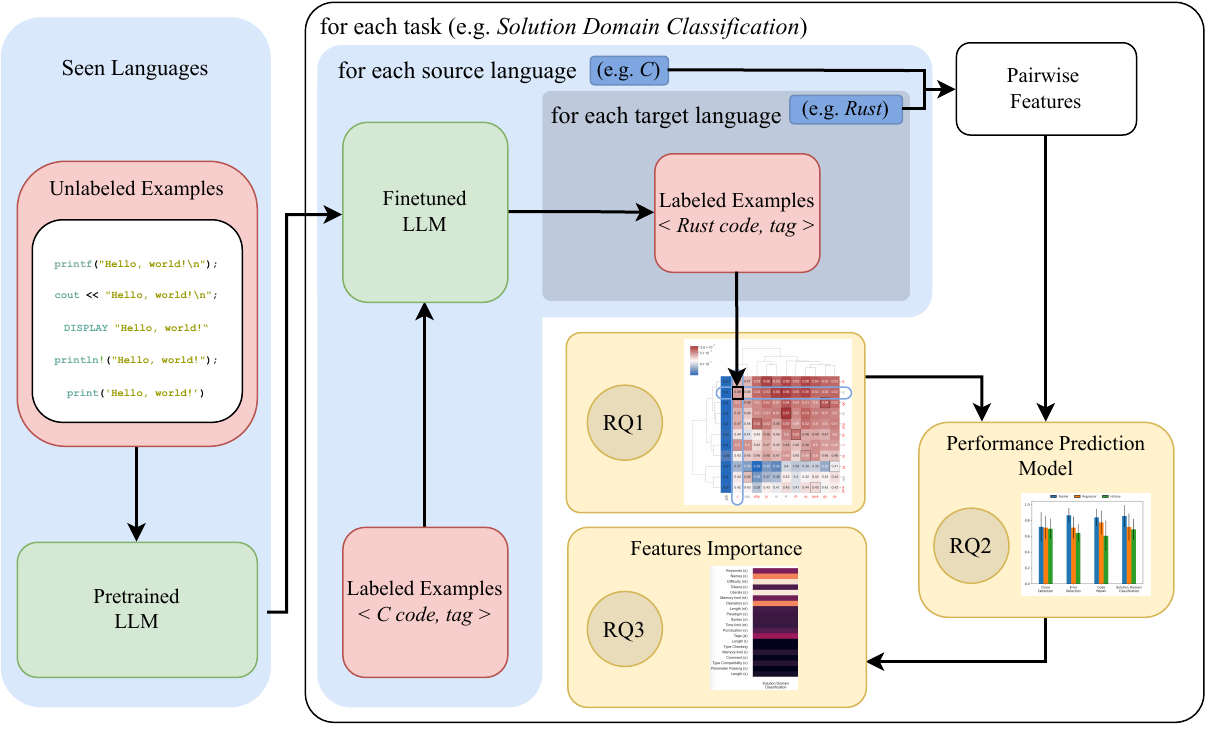}}
  \caption{Overview. Consider an LLM pretrained on unlabeled code in multiple seen languages. Finetune on task-specific labeled samples from a source language. For RQ1, test performance on a target language. For RQ2, train a model that ranks transfer performance given features of a language pair. For RQ3, measure how important the features of language pairs are. Repeat for several languages and four tasks.}
  \label{fig:teaser}
\end{figure*}

Figure~\ref{fig:teaser} summarizes the methodology
for the finetuning part of our empirical study, which builds upon the work of~\citet{de2022make} by exploring programming languages and multiple tasks.
On the left, we start with a pretrained LLM that has seen
unlabeled code in several programming languages.
Next, for each of four tasks, for each source language with a suitable
amount of labeled training data for that task, we finetune the LLM.
For each target language with a suitable amount of labeled test data
for the task, we evaluate the finetuned LLM. 
For example, Figure~\ref{fig:teaser} illustrates this with a
\tagtasklower task, using C as the source language and Rust as
the target language, which yields an F1 score of~$0.48$.
Doing this for four different tasks and all their source and target
language combinations yields four different heat maps.
The methodology using in-context learning on the fifth task is
analogous and yields a fifth heat map.

This paper addresses three research questions (RQ1--RQ3).

\textbf{RQ1:}~\textit{How well does cross-lingual transfer work for a
  given task across different language pairs?}
We answer this question directly by exploring the heat maps.

\textbf{RQ2:}~\textit{Given a task and target language, how should we
  pick the source language for best performance?}
We answer this question by training a performance prediction model
from the language pair features and the ground-truth target labels
from the heat maps, which predicts the ranking of source languages.
For example, in Figure~\ref{fig:teaser}, the performance prediction
model for the \tagtasklower task might rank C behind Kotlin as a source
language for transfer to Rust.
Given the cost of training {an LLM}, the performance prediction model is
useful in directing data acquisition efforts and deciding how to spend
compute resources.

\textbf{RQ3:}~\textit{Which characteristics of a language pair are
  predictive of transfer performance, and how does that depend on the
  given task?}
We answer this question by measuring the importance of language pair
features in the prediction model from the previous experiment.

\noindent
\\This paper makes the following contributions:
\begin{itemize}
\item Evaluating the pairwise transferability for several programming languages (including low-resource ones) and five tasks using language models.
\item Developing a method to predict the best language to transfer from, for different target languages and tasks.
\item Characterizing the features of programming language pairs that are predictive of transferability for given tasks.
\end{itemize}

\noindent
One goal of this paper is to offer practical guidance to engineers
who build LLM-driven software engineering tools towards
more informed choices for data acquisition and modeling.
A second goal of this paper is to 
use the lens of transfer learning to shed light on
programming language characteristics.
Some highlights of our results are as follows:
\begin{itemize}
    \item Learning transfers well for all tasks and cross-lingual learning transfers better than zero-shot.
    \item Transfer learning depends on the target language. Java and Go are the best target languages whereas C++ and Python are the worst.
    \item Transfer learning also depends on source language used for finetuning. Kotlin and JavaScript are the best source languages, C++ is the worst.
    \item We can reliably predict which source languages will perform well on a given task by considering linguistic, synthetic, dataset-specific, and model-specific features.
    \item Different tasks rely on different features;
      for some, keywords and names are most important.
\end{itemize}

\noindent
Assuming software engineers increasingly benefit from LLM support, we
believe this paper will help democratize such support into low-resource settings {and help inspire more research in this area}.

\section{Related Work}
\label{sec:related}

\paragraph{Cross-lingual transfer for natural languages.}
\citet{lin_et_al_2019} use 4~tasks to study transfer among up to 60
natural languages and explore how important features of language pairs
are for predicting how well learning transfers between them, finding
that feature importances vary a lot across tasks.
\citet{lauscher_et_al_2020} use 5~tasks to study transfer among up to
15~languages and argue that while zero-shot performance can work for
low-level tasks, higher-level tasks benefit from at least a few
target-language shots.
\citet{de2022make} use 1~task (POS tagging) with 65~source and
105~target languages and study the effects of language families and
writing systems, with Romanian emerging as a particularly good source
language.
\citet{ahuja_et_al_2022} use 11~tasks with 1~source and varying numbers
of target languages, confirming earlier findings that feature
importances vary a lot across tasks.
Our paper takes inspiration from these works but differs by studying
programming languages instead of natural languages, which have
different tasks and different features affecting transferability.

\paragraph{Cross-lingual transfer for programming languages.}
\citet{zhou2022improving} call for studying transfer, motivating with
1~task (code completion) and 2~languages (from Hack to Rust).
\citet{chen2022transferability} use 2~tasks (code summarization and
search) and 6~languages to study transfer to Ruby, and propose
picking the source language based on language similarity.
\citet{ahmed2022multilingual} use 3~tasks to study transfer among
6~languages, demonstrating that due to the nature of their tasks,
signals from identifiers are highly important for transferability.
\citet{yuan_et_al_2022} use 1~task (automated program repair) to study
transfer among 5~languages, sequentially fine-tuning on multiple
languages with innovative tricks to prevent catastrophic forgetting.
\citet{pian_et_al_2023} use 2~tasks (code summarization and
completion) to study transfer among 4~languages, using meta-learning
to improve a base learner.
\citet{cassano_et_al_2024-oopsla} introduce MultiPL-T, a new
semi-synthetic training dataset, and show that it improves transfer
for 1~task (NL to code completion) across 19 languages. Our paper also focuses on programming languages, but considers more
tasks and more languages to explore conditions for effective transfer.

There is also work on learning across multiple programming languages that has been instrumental in making studies like ours possible, but it differs in that it does not focus on cross-lingual transfer.
Unsupervised pretraining on multiple languages has 
become common, but transferability of supervised tasks has not yet been
thoroughly studied; our paper addresses that gap.
One could also imagine exploring transfer from multiple source languages
to a low-resource target; we leave this to future work.

\section{Experimental Setup}
\label{sec: approach}

\begin{figure}[!t]
  \centering
      \begin{subfigure}{\textwidth}
        \centerline{\includegraphics[width=1\linewidth]{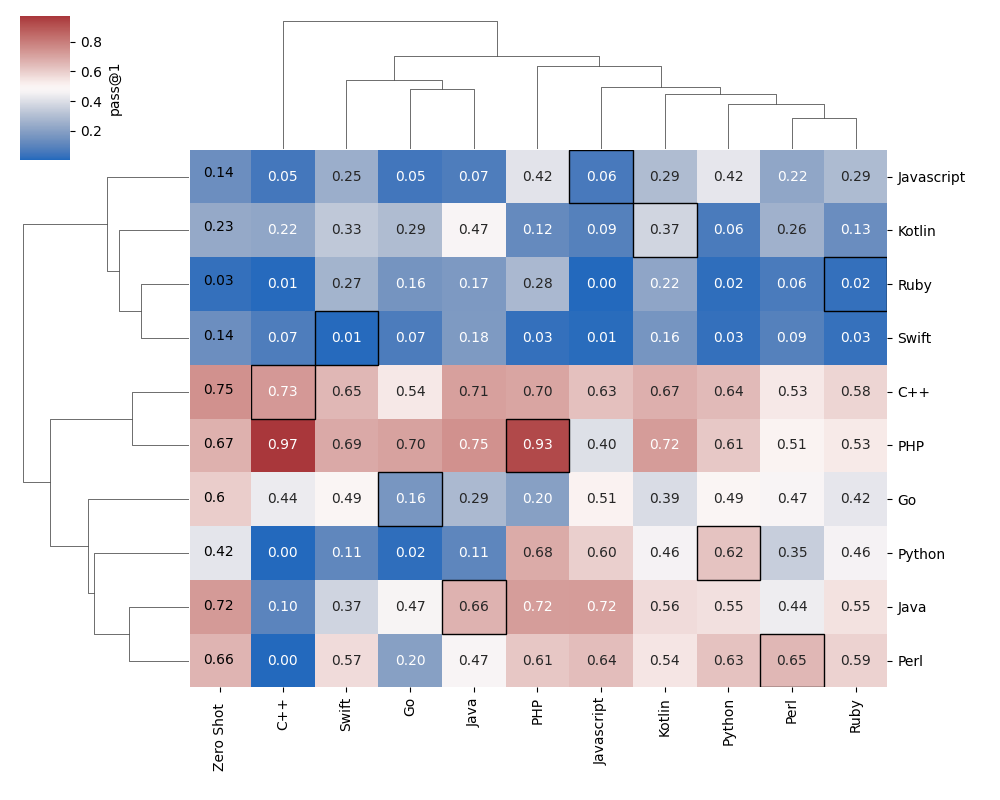}}
        \vspace*{-2mm}
        \subcaption*{{Code Generation: 10 source languages $\times$ 10 target languages. Metric: Pass@1 Score.}}
    \end{subfigure}

  \caption{{Few shot prompting transfer scores heat map. The figure shows scores for every combination of source and target language. Each column corresponds to a source language, with ``Zero Shot'' showing zero shot performance. Each row corresponds to a target language. Similarly, dendrograms show results of hierarchical clustering and framed black boxes highlight the performance of a source language on itself as the target language.}}
  \label{fig:zsh_mbxp}
\end{figure}

Most of our experiments are based on finetuning a small model for four tasks as illustrated in Figure~\ref{fig:teaser}; in addition, there are also some experiments based on few-shot prompting a larger model without finetuning.
For most experiments, given a task, our experimental
approach first finetunes the CodeT5~\citep{wang2021codet5} model for each source language individually
and then tests each finetuned model on all target languages.
We applied this approach to four tasks where the number of source languages varies from 6 to 22, leading to 58 finetuned models in total. For each source language, we finetune the model using one A100 GPU for 6~to~30 hours depending on the task. 
Each model was then evaluated on 11 to 43 target languages producing 1,808 experiments. All results are presented in \mbox{Figures~\ref{fig:transfer_scores}(a)--(d)} and analyzed in Section~\ref{sec:results}.
For the few-shot prompting experiments, we generated responses with the \LlamaInstr model via~\citealp{togetherai} using a temperature of 0.8. Those results are presented in Figure~\ref{fig:zsh_mbxp}.

\begin{figure*}[!t]
    \centering
    \begin{subfigure}{\textwidth}
        \centerline{\includegraphics[width=1\linewidth]{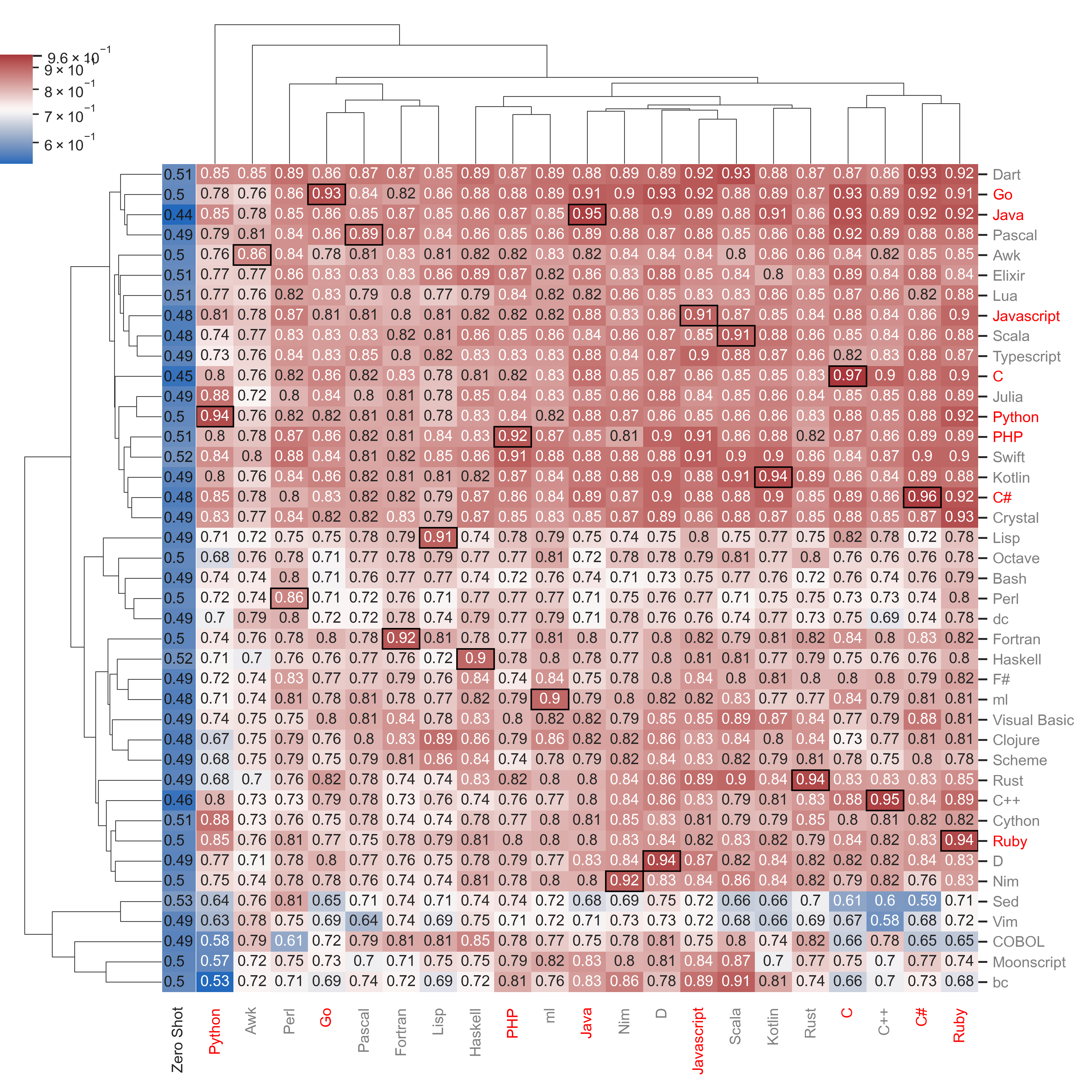}}
        \vspace*{-2mm}
        \subcaption{\clonetask: 21 source languages $\times$ 41 target languages. Metric: F1 Score.}
        \label{fig:zsh_clone}
    \end{subfigure}
    \vspace*{-2mm}
    \caption{Transfer scores heat map.
    The figure shows scores for every combination of source and target language.
    Each column corresponds to a source language, with ``Zero Shot'' showing zero shot performance, i.e., without fine-tuning on any source language.
    Each row corresponds to a target language.
    The languages whose language-name label uses red font were seen during pre-training.
    Framed black boxes highlight the performance of a source language (column) on itself as the target language (row).
    The dendrograms show results of hierarchical clustering based on similarity of the performance vectors.
    The row and column order is also determined by the same clustering. {}}
    \vspace*{-2mm}
\end{figure*}

\begin{figure*}[!t]\ContinuedFloat
    \begin{subfigure}{\textwidth}
        \centerline{\includegraphics[width=1\linewidth]{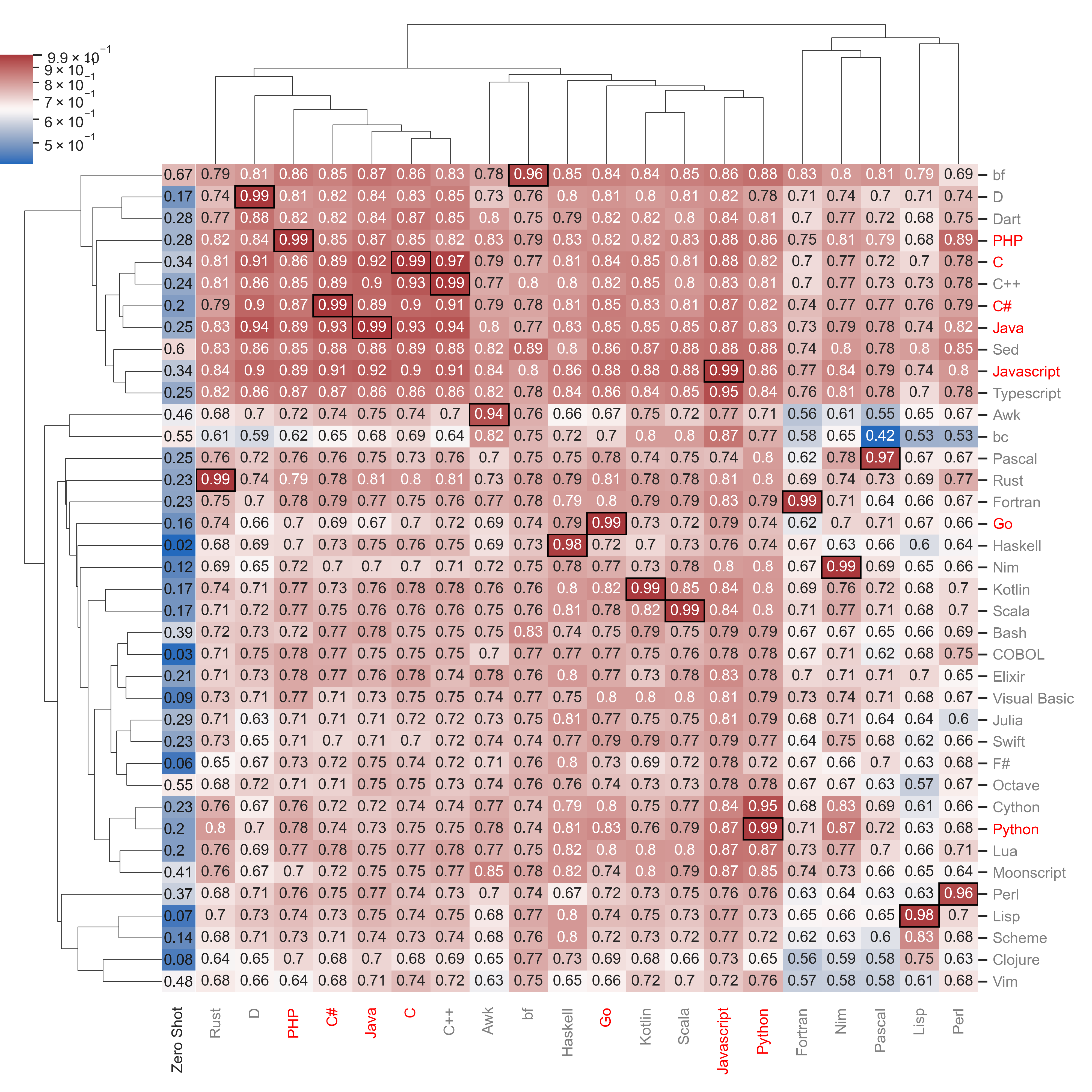}}
        \vspace*{-2mm}
        \subcaption{\refinetask: 20 source languages $\times$ 38 target languages. Metric: BLEU score.}
        \label{fig:zsh_refine}
    \end{subfigure}
    \vspace*{-2mm}
    \caption{(continued; see caption from \ref{fig:zsh_clone})}
    \vspace*{-2mm}
\end{figure*}

\begin{figure*}[!t]\ContinuedFloat
    \begin{subfigure}{\columnwidth}
        \centerline{\includegraphics[width=0.75\linewidth]{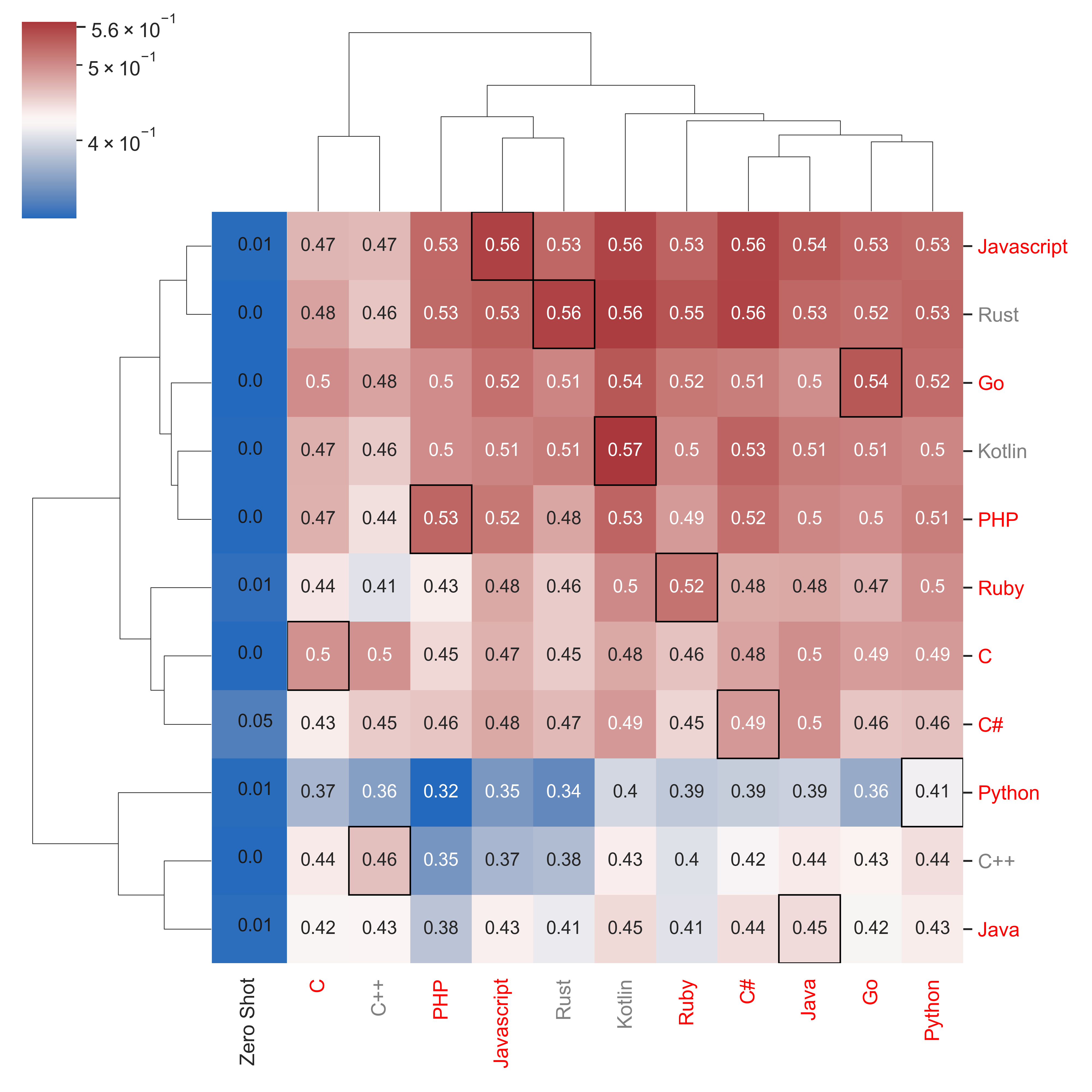}}
        \vspace*{-2mm}
        \subcaption{\tagtask: 11 source languages $\times$ 11 target languages. Metric: F1 score.}
        \label{fig:zsh_tag}
    \end{subfigure}
    \vspace*{-2mm}
    \caption{(continued; see caption from \ref{fig:zsh_clone})}
    \vspace*{-2mm}
\end{figure*}

\begin{figure*}[!t]\ContinuedFloat
    \begin{subfigure}{\columnwidth}
        \centerline{\includegraphics[width=0.7\linewidth]{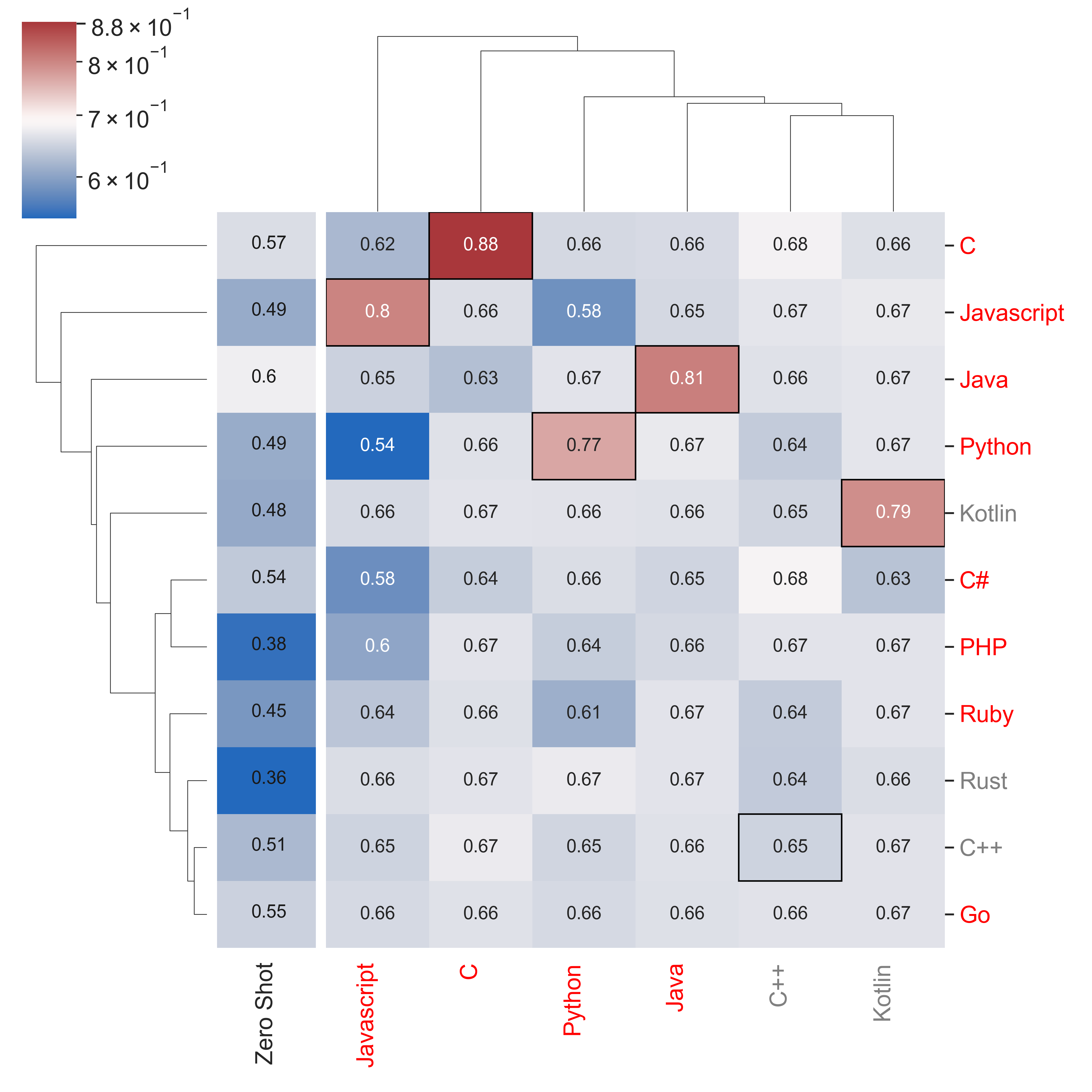}}
        \vspace*{-2mm}
        \subcaption{\compiletask: 6 source languages $\times$ 11 target languages. Metric: F1 Score.}
        \label{fig:zsh_comp}
    \end{subfigure}
    \vspace*{-2mm}
    \caption{(continued; see caption from \ref{fig:zsh_clone})}
    \vspace*{-2mm}
    \label{fig:transfer_scores}
\end{figure*}

\paragraph{Base datasets.} 
The {five} tasks studied are derived from {three} multilingual code datasets: CodeNet, XCodeEval,  {and MBXP}.
CodeNet~\citep{puri2021codenet} consists of about 14M code samples
in 55 different programming languages, derived from submissions to online coding judge websites.
The dataset comes with benchmarks for code classification and code similarity.
XCodeEval~\citep{khan_et_al_2023} contains 25M samples in 11 programming languages from another different online coding judge website.
This dataset comes with an execution-based evaluation framework and several different benchmarks: from classification (\tagtasklower and \compiletasklower) to generation (program synthesis, automatic program repair, and code translation). 
{MBXP~\citep{athiwaratkun2023multilingual} is a multilingual extension of MBPP \citep{austin2021program} to Java, JavaScript, TypeScript, Go, Ruby, Kotlin, PHP, C\#, Scala, C++, Swift, and Perl. A companion code package is provided to perform execution in all supported languages. The dataset is designed for investigating the effectiveness of few-shot prompting,  zero-shot translation, as well as the robustness to prompt perturbation. In our experiment, we utilize the few shot prompting setup.}

\paragraph{Tasks.}

To study how learning transfers between programming languages, we explore two type of tasks: classification~(\compiletasklower, \tagtasklower, \clonetasklower) and generation (\refinetasklower, code generation).
For the classification tasks, we follow \citet{lewis-etal-2020-bart} by predicting
class labels based on the final decoder hidden state.
We use F1 score for evaluating the classification tasks, which is a widely accepted practice since it is good at dealing with class imbalance.
We use BLEU score for evaluating the \refinetasklower task; while an execution-based metric would provide value, it was infeasible due to the large number of languages in CodeNet~\citep{puri2021codenet}. (Other prominent benchmarks for the \refinetasklower task also adopt BLEU score~\citep{lu2021codexglue}.)
Finally, we use execution-based Pass@1 score for evaluating the code generation task.
The execution-based evaluation checks the functional correctness of the generated code by running it against the provided tests.

\textit{\compiletask:}
Given a code $C$ in language $L$, do binary classification of whether $C$ compiles (or can be loaded by an interpreter) without error.

\textit{\tagtask:}
Given a code $C$ in language~$L$, do multi-label classification into a set of tags corresponding to algorithmic techniques required to write the solution (e.g., sorting, graphs). 

\textit{\clonetask:}
Given two code samples $C_1$ and $C_2$ in language $L$, do binary classification of whether they are \mbox{type-IV} clones~(semantically similar)~\citep{roy_cordy_2007}. We derived the dataset from CodeNet as follows. Given all combinations of solutions to all problems in language $L$, we identify positive samples (clones) as pairs of accepted solutions for the same problem and the others as negative examples. To balance the data across problems,  we ensure a ratio of 0.15 of positive samples across different languages.

\textit{\refinetask:}
Given a buggy code~$C$ in language~$L$, generate the corresponding repaired code.
We derived the dataset from CodeNet, modifying samples by sequentially
inserting, removing, or replacing tokens of different types for fixed
ratios for different token types.

\textit{Code Generation:} Given a prompt $P$ consisting of a function signature and a docstring in language $L$, the task is to generate the function body completion. In the few shot prompting setup, few-shot prompts consisting of three correct functions from the respective MBXP dataset are provided. Few shot examples precede the function completion prompt for each evaluation.

\paragraph{Data sampling.}
Our objective is to facilitate an extensive study on transfer across the maximum 
number of language pairs, which can be helpful for those working with low-resource 
languages. 
Sources are selected based on the relative performance of fine-tuned checkpoints 
compared to a random baseline, in order to determine if there is enough signal in 
the data to facilitate learning. 
Languages that demonstrate higher performance than the baseline are included as 
sources. Given variation in the size of datasets for
different languages, we follow the sampling procedure of 
\citet{de2022make} to select the number of 
training samples and identify potential source or high-resource languages.
We first finetune the model with datasets of different 
number of samples $N$=\{10\textrm{K}, 30\textrm{K}, 50\textrm{K}, 70\textrm{K},  100\textrm{K}\}. 
Depending on the number of training examples $N_L$ 
for a language~$L$, we randomly upsample languages with 
We filter the set of source languages for a given task based on the 
relative performance compared to a baseline model. 
We select sample sizes of 50K for \tagtasklower, 70K for \compiletasklower, and 100K for \clonetasklower and \refinetasklower.

\paragraph{Models.}

Our finetuning experiments are based on CodeT5-base (220M parameters) using the Hugging Face transformers library \citep{wolf2020huggingfaces}. 
Code-T5 is an open-source model, pre-trained on eight 
programming languages
(Ruby, JavaScript, Go, Python, Java, PHP, C, and C\#). 
We consider these languages as \emph{seen languages} and 
Figure~\ref{fig:transfer_scores} highlights the names of these pre-training languages in red font.
Due to its encoder-decoder nature, the model performs well 
on both code generation and code understanding 
tasks~\citep{wang2021codet5} like \refinetasklower, \compiletasklower,
and \clonetasklower, among others.
This, along with its relatively small size, makes it a good 
fit for our empirical study, which requires
58 finetuned models with numerous inference runs each.
{We keep the same hyperparameters for all the experiments: learning rate of 2e-5, batch size of 8, and the number of epochs set to 20.} {The few-shot prompting experiment use \LlamaInstr,\footnote{{\url{https://huggingface.co/meta-llama/Llama-3.3-70B-Instruct}}} an instruction-tuned multilingual LLM (70B parameters) \citep{grattafiori2024llama3herdmodels}. }

\paragraph{Data Splits}

\begin{table*}
\centering
\begin{tabular}{lrr}
\toprule
Language & \multicolumn{1}{c}{Train} & \multicolumn{1}{c}{Test}\\
\cmidrule(lr){1-1}\cmidrule(lr){2-3}
  php    & 914,924 & 1,975\\
  rb     & 912,159 & 3,963\\
  hs     & 843,011 & 2,045\\
  py     & 785,760 & 4,235\\
  kt     & 731,575 & 1,552\\
  js     & 713,854 & 1,600\\
  c      & 555,924 & 2,363\\
  scala  & 473,594 & 1,705\\
  go     & 352,181 &   867\\
  java   & 266,248 & 1,338\\
  d      & 247,954 & 1,355\\
  cs     & 233,059 & 1,151\\
  pl     & 185,685 & 1,762\\
  nim    & 175,839 &   765\\
\bottomrule
\end{tabular}
\hspace*{5mm}
\begin{tabular}{lrr}
\toprule
Language & \multicolumn{1}{c}{Train} & \multicolumn{1}{c}{Test}\\
\cmidrule(lr){1-1}\cmidrule(lr){2-3}
  ml     & 175,056 & 2,008\\
  f      & 172,046 & 1,253\\
  cpp    & 150,492 & 1,792\\
  pas    & 136,961 & 1,370\\
  rs     & 128,271 &   478\\
  awk    &  95,312 & 1,035\\
  jl     &  88,700 & 1,737\\
  sh     &  64,899 & 1,070\\
  ts     &  47,414 &   799\\
  l      &  45,986 &   585\\
  swift  &  38,531 &   737\\
  fs     &  31,001 &   962\\
  scm    &  22,787 &   826\\
  cr     &  21,733 & 1,356\\
\bottomrule
\end{tabular}
\hspace*{5mm}
\begin{tabular}{lrr}
\toprule
Language & \multicolumn{1}{c}{Train} & \multicolumn{1}{c}{Test}\\
\cmidrule(lr){1-1}\cmidrule(lr){2-3}
  sed    &  21,584 & 1,349\\
  lua    &  20,688 & 1,253\\
  pyx    &  18,184 &   931\\
  dc     &  17,322 & 1,706\\
  vb     &   7,003 & 1,132\\
  octave &   5,370 & 1,476\\
  vim    &   4,176 &   745\\
  cob    &   3,437 & 1,038\\
  clj    &   1,058 &   910\\
  dart   &     506 &   480\\
  moon   &     242 &   453\\
  ex     &     330 &   549\\
  \\
  \\
\bottomrule
\end{tabular}
\caption{\label{table:clone_data_sample_count}Number of train and test samples for \clonetask for each language.}
\end{table*}

Train and test splits for \compiletask and \tagtask are provided from the original dataset as described in the dataset statistics from 
xCodeEval~\citep{khan_et_al_2023}. For Code Repair, 50,000 training examples and 1,000 test examples are synthetically generated for each language.  For \clonetask, a distinct set of problems is used for train and test splits, including languages with a minimum of 450 test examples. Table~\ref{table:clone_data_sample_count} shows the number of examples for \clonetask for each language. {In the few-shot prompting experiment, we conduct a cross-lingual evaluation across all language pairs, using nearly 1,000 examples for each. We additionally use the provided synthetic canonical solutions to measure dataset features across overlapping problem solutions.}

\section{Results and Discussion}

\label{sec:results}

This section addresses the research questions RQ1--RQ3 stated in the
introduction.

\subsection{Transfer analysis}
\label{transfer_analysis}
We explore RQ1 via extensive experiments on every combination of source and target programming language, for each task.
The four heat maps of Figure~\ref{fig:transfer_scores} show the results.
A \emph{source} programming language is the language used for fine-tuning a model, with samples in the training data.
A \emph{target} programming language is the language used to evaluate a model, with samples in the test data.
Figure~\ref{fig:transfer_scores} shows source languages on the horizontal axis and target languages on the vertical axis.
Monolingual scores (where the source and the target language are the same) are highlighted with bold boxes.
The first column from the left indicates zero-shot 
performance of the base Code-T5 model on different target languages.
For RQ1 we analyze model performance trends with respect to task, 
source language, and target language.

\paragraph{Task dependency.}
The heat maps in Figure~\ref{fig:transfer_scores} show that transferability of source languages varies depending on the task. 
Visually this can be seen by how the color gradient varies for each task.
To quantify this observation, 
we calculate the mean scores across all the language combinations.
The mean cross-lingual scores are 0.78 for 
\clonetasklower, 0.75 for \refinetasklower, 0.65 for \compiletasklower, and 0.47 for \tagtasklower.
The monolingual scores~(in bold boxes) are higher~(more red) than
the cross-lingual scores, averaging
0.91, 0.98, 0.78, and 0.51 for the same four tasks.
The zero-shot columns, where the model is not finetuned on any source
language, have lower scores (more blue) for all heat maps,
averaging 0.49, 0.28, 0.49, and 0.01.
Variations in the mean monolingual, cross-lingual, and zero-shot scores 
across tasks confirm that model performance is task dependent.
The difference in overall zero-shot and cross-lingual scores indicates 
that in the absence of fine-tuning data for any task in any programming language,
it is better to finetune with some other language in that task 
than to use a zero-shot setting.
This cross-lingual advantage holds for every task.

\begin{table*} 
  \centerline{
  \begin{tabular}{lcccc}
  \toprule
  & \multicolumn{4}{c}{Mean Score} \\
  \cmidrule(lr){2-5}
  Task & Mono & Cross & Overall & Zero-Shot\\
  \cmidrule(lr){1-1}\cmidrule(lr){2-5}
  \clonetask     & 0.91  & 0.78     & 0.79    & 0.49  \\ 
  \refinetask    & 0.98  & 0.75     & 0.76    & 0.28  \\ 
  \compiletask   & 0.78  & 0.65     & 0.66    & 0.49  \\
  \tagtask       & 0.51  & 0.47     & 0.47    & 0.01 \\
  \bottomrule
  \end{tabular}
  }
  \caption{\label{table:scores}Mean Scores by Task. Monolingual implies finetuning where the train and test data language is the same. Cross-lingual implies finetuning where the train and test data languages differ. Overall scores include both monolingual and cross-lingual scenarios. Zero-shot means the performance of the base pre-trained model (CodeT5-220M) on the test set without finetuning.}
  \end{table*}

Looking at these scores in absolute terms reveals that the choice of tasks 
also affects transferability.
For \tagtasklower, the difference in mean cross-lingual and monolingual scores 
is only 0.04, whereas for \refinetasklower, it is 0.23.
The difference in mean cross-lingual and zero-shot scores for the same two tasks is 0.46 
and 0.48.
Although overall trends show learning transfers well in all tasks, cross-lingual 
performance in relation to monolingual performance is highly task 
dependent.
Within each task, model performance depends on the choice 
of source and target programming language, which we explore next. 
Table~\ref{table:scores} shows the mean score by task.

\begin{tcolorbox}[boxsep=0mm]
  Learning transfers well for all tasks.
  Cross-lingual learning works better than zero-shot.
\end{tcolorbox}\vspace*{-3mm}

\paragraph{Target language dependency.}

\begin{table*}
  \centerline{\begin{tabular}{l cc cc cc cc c}
  \toprule

   & \multicolumn{2}{c}{\clonetask} & \multicolumn{2}{c}{\refinetask} & \multicolumn{2}{c}{\compiletask} & \multicolumn{2}{c}{Solution Domain} &  \\
    \cmidrule(lr){2-3}\cmidrule(lr){4-5}\cmidrule(lr){6-7}\cmidrule(lr){8-9}
  Target & cross & 0-shot    & cross   & 0-shot & cross     & 0-shot & cross & 0-shot & \multicolumn{1}{c}{Mean Rank} \\ 
  \cmidrule(lr){1-1}\cmidrule(lr){2-3}\cmidrule(lr){4-5}\cmidrule(lr){6-7}\cmidrule(lr){8-9}\cmidrule(lr){10-10}
  Java       & {0.86} & 0.44 &  {0.84} & 0.25 &  {0.66}  & 0.60 & {0.42} & 0.009 & 4.25     \\ 
  Go       & {0.86} & 0.50 & {0.71} & 0.16 &  {0.66} & 0.55 &  {0.51} & 0.003 & 4.25     \\ 
  Rust       & {0.80}  & 0.49 & {0.77} & 0.23 & {0.66}  & 0.36 & {0.53} & 0.001  & 4.50    \\
  JavaScript       &  {0.83} & 0.48 &  {0.85} & 0.34 &  {0.65}  & 0.49 & {0.52} & 0.008 & 4.75     \\ 
  Kotlin       &  {0.84}  & 0.49 &  {0.76} & 0.17 &  {0.66}  & 0.48 & {0.50} & 0.005 & 5.00     \\ 
  PHP       & {0.84}  & 0.51 & {0.82} & 0.28 & {0.65}  & 0.38 & {0.50} & 0.004 & 5.25    \\ 
  C\#       &  {0.84}  & 0.48 & {0.82} & 0.2 & {0.64} & 0.54 &  {0.47} & 0.046 & 5.50    \\ 
  C       &  {0.83} & 0.45 & {0.82} &  0.34 &  {0.66} & 0.57 & {0.48} & 0.004 &  5.75    \\ 
  C++       &   {0.79}  & 0.46 &  {0.81} & 0.24 &  {0.66} & 0.51 &  {0.41} & 0.004 & 7.00    \\ 
  Python    & {0.83}  & 0.50 & {0.76} &  0.20 & {0.64} & 0.49 &  {0.37} & 0.006 & 8.75    \\
  \bottomrule
  \end{tabular}}
  \caption{\label{table:target_transfer_scores}Score distribution by target languages common for all tasks. Scores are the mean of the score of every source for a given target language. Mean rank is used to rank the languages based on their ranking for each task. All the languages shown are high-resource target languages for \clonetask, \refinetask and \tagtask, which means they are also among the source languages for those tasks. C\#, Go, PHP, Ruby, and Rust are low-resource languages for \compiletask. The table shows that all languages benefit from transfer learning. Java and Go seem to benefit the most and C++ and Python seem to benefit the least.}
  \end{table*}

In Figure~\ref{fig:transfer_scores}, high-resource target programming 
languages are represented on both axes.
Low-resource target programming languages are represented only on 
the vertical axis.
Models are finetuned only on high-resource programming languages 
and tested on both high- and low-resource programming languages.

To understand overall trends across all four tasks, we consider ten target programming languages that are common to 
all four tasks. 
Six (Kotlin, JavaScript, Java, Python, C, and C++)
are high-resource in all four tasks.
Four (Go, Rust, PHP, and C\#) are low-resource in \compiletasklower.
For each of these ten languages, for each of the four tasks, we calculate 
the mean cross-lingual score using all scores in the row of 
Figure~\ref{fig:transfer_scores} for each language, excluding the monolingual  
score in the bold box and the zero-shot score in the extreme left.
Based on this mean we rank each of the 10 languages for each task, 
and then calculate the mean rank for each language. 
The most transferrable target languages from best to worst are 
Java, Go, Rust, JavaScript, Kotlin, PHP, C\#, C, C++, and Python.
Some target programming languages are good at monolingual performance, but much worse at cross-lingual performance.
For example,
in-language performance for C++ is 0.95, 0.99, and 0.46 for \clonetasklower, \refinetasklower, and \tagtasklower, whereas its mean cross-language performance is 0.79, 0.81, and 0.41 for the same tasks. 
This might be explained by the fact that C++ has a reputation as being
one of the hardest programming languages to learn for human software
developers.
All numbers are in Table~\ref{table:target_transfer_scores}.

\begin{table*}
  \centerline{\begin{tabular}{lllllc}
  
  \toprule
  \multirow{2}{*}{} & \multicolumn{2}{c}{\clonetask} & \multicolumn{2}{c}{\refinetask}  &  \\
  \cmidrule(lr){2-3}\cmidrule(lr){4-5}
  Target Language & cross & 0-shot & \multicolumn{1}{c}{cross} & 0-shot & Mean Rank \\ 
  \cmidrule(lr){1-1}\cmidrule(lr){2-3}\cmidrule(lr){4-5}\cmidrule(lr){6-6}
  
  Dart       & {0.87} & 0.51 &  {0.79} & 0.28 &  2.0     \\ 
  
  TypeScript       & {0.83} & 0.49 &  {0.83} & 0.25 &  3.5     \\ 
  
  Elixir       & {0.83} & 0.51 &  {0.75} & 0.21 & 5.0     \\ 
  
  Lua       & {0.82}  & 0.51 & {0.77} & 0.20 & 5.0     \\ 
  
  Swift       & {0.85}  & 0.52 & {0.72} & 0.23 & 6.5     \\ 
  
  Visual Basic       & {0.80} & 0.49 &  {0.75} & 0.09 &  7.5     \\ 
  
  Cython       & {0.79} & 0.51 &  {0.75} & 0.23 & 7.5     \\ 
  
  Julia       & {0.83} & 0.49 &  {0.72} & 0.29 & 8.0     \\ 
  
  Moonscript       & {0.74} & 0.50 &  {0.75} & 0.41 & 9.0     \\ 
  
  Sed       & {0.69}  & 0.53 & {0.85} & 0.60 & 9.5     \\ 
  
  Scheme       & {0.78}  & 0.49 & {0.71} & 0.14 & 11.5     \\ 
  
  Shell       & {0.74}  & 0.49 & {0.74} & 0.39 & 12.0     \\ 
  
  Clojure       & {0.80}  & 0.48 & {0.67} & 0.08 & 12.5     \\ 
  
  COBOL       & {0.73}  & 0.49 & {0.74} & 0.03 & 12.5     \\ 
  
  F\#       & {0.78}  & 0.49 & {0.71} & 0.06 & 13.0     \\ 
  
  Octave       & {0.76}  & 0.50 & {0.71} & 0.55 & 13.0     \\ 
  
  bc       & {0.73}  & 0.50 & {0.67} & 0.55 & 16.5     \\ 
  
  Vim       & {0.69}  & 0.49 & {0.67} & 0.48 & 16.5     \\ 
  
  \bottomrule
  \end{tabular}}
  \caption{\label{table:lr_transfer_scores}Score distribution for low resource target languages common for \clonetask and \refinetask. Scores are the mean of the score of every source language for the given target language. Mean rank is used to rank the languages based on their ranking across each task. All languages benefit from transfer learning. Dart, TypeScript benefit the most and bc, Vim benefit the least.}
  \end{table*}

We do a similar analysis for low-resource target languages in \clonetasklower 
and \refinetasklower, since these two tasks have a wider and more 
varied set of low-resource languages.
Here, the most transferrable target languages from best to worst are 
Dart, TypeScript, Elixir, Lua, Swift, COBOL, F\#, Octave, bc, and Vim.
See Table~\ref{table:lr_transfer_scores} for details.

For \compiletasklower, only a smaller set of languages can be considered low resource. 
Average scores vary less across different 
sources with highest being Rust with 0.66 and lowest being 
C\# with 0.64. 
For \tagtasklower there are no low-resource target languages.

\begin{tcolorbox}[boxsep=0mm]
  While all target languages benefit from cross-lingual training,
  some benefit much more than others.
  Among low-resource target languages, Dart and TypeScript benefit the most.
\end{tcolorbox}\vspace*{-3mm}

\paragraph{Most Transferable Source Language.}

\begin{table*}
  \centerline{\begin{tabular}{l cc cc cc cc c}
  \toprule
  
  & \multicolumn{2}{c}{\clonetask} & \multicolumn{2}{c}{\refinetask} & \multicolumn{2}{c}{Solution Domain} & \multicolumn{2}{c}{\compiletask} &  \\
  \cmidrule(lr){2-3}\cmidrule(lr){4-5}\cmidrule(lr){6-7}\cmidrule(lr){8-9}
  Source & score & rank & score & rank  & score & rank  & score & rank & Mean Rank \\ 
  \cmidrule(lr){1-1}\cmidrule(lr){2-3}\cmidrule(lr){4-5}\cmidrule(lr){6-7}\cmidrule(lr){8-9}\cmidrule(lr){10-10}
  
  Kotlin       & {0.8}   & 3    & {0.78}  & 3  & {0.5}  & 1   & {0.67}  & 1 & 2.0 \\ 
  
  JavaScript       & {0.82}   & 1    & {0.81}  & 1  & {0.47}  & 4   & {0.62}  & 6 & 3.0 \\ 
  
  Java       & {0.8}   & 4    & {0.78}  & 5  & {0.49}  & 3   & {0.66}  & 2 & 3.5 \\ 
  
  Python       & {0.73}   & 6    & {0.79}  & 2  & {0.49}  & 2   & {0.66}  & 5 & 3.75 \\ 
  
  C       & {0.8}   & 2    & {0.78}  & 4  & {0.45}  & 5   & {0.66}  & 4 & 3.75 \\ 
  
  C++       & {0.79}   & 5    & {0.78}  & 6  & {0.44}  & 6   & {0.66}  & 3 & 5 \\ 
  
  \bottomrule
  \end{tabular}}
  \caption{\label{table:source_ranking} Source languages common for all tasks ranked by mean score across target languages and tasks. Score is the mean of the scores of all target languages for a given source language. We see that Kotlin is relatively the best source language and C++ the worst. This result is surprising since Kotlin is not a seen language during pre-training.}
  \end{table*}

The number of source languages vary for each of the four tasks.
To understand the source language dependency and to identify 
the most transferrable source language, we start from languages 
all four tasks have in common
(C, C++, Java, JavaScript, Kotlin, and Python).
For each task, for each language, we calulate the mean 
cross-lingual score.
Once we have the cross-lingual mean for each language, we rank the 
source languages based on their mean score for each task.
Based on these rankings, we calculate the mean rank, for overall cross-lingual performance of source languages, across all 
target languages, across all tasks.

The most transferrable source languages from best to worst are 
Kotlin, JavaScript, Java, Python, C, and C++.
It may be surprising that Kotlin is the best-performing source
language, since it is rarely discussed in the context of training LLMs 
and is not part of the pre-training corpus for CodeT5~\citep{wang2021codet5}.
Prior work on transfer learning for natural
languages~\citep{de2022make} yielded a similar surprise, with Romanian
as the best-performing source language.
A possible explanation could be that Romanian and Kotlin both have
strong roots in dominant languages of the past (Latin and Java), while
sitting at a cross-roads that helped them absorb many influences.
Detailed scores and rankings are in Table~\ref{table:source_ranking}.

\begin{table*}
  \centering
  \begin{tabular}{lccccccc}
  \toprule
   & \multicolumn{2}{c}{\clonetask} & \multicolumn{2}{c}{\refinetask} & \multicolumn{2}{c}{\compiletask} &  \\
  \cmidrule(lr){2-3}\cmidrule(lr){4-5}\cmidrule(lr){6-7}
   Source & \multicolumn{1}{c}{score}   & rank    & \multicolumn{1}{c}{score}     & rank  & \multicolumn{1}{c}{score} & rank & Mean Rank \\ 
   \cmidrule(lr){1-1}\cmidrule(lr){2-3}\cmidrule(lr){4-5}\cmidrule(lr){6-7}\cmidrule(lr){8-8}
  
  JavaScript       & {0.81}   & 1    & {0.82}  & 1  & {0.63}  & 6 & 2.67 \\ 
  
  Java       & {0.79}   & 2    & {0.75}  & 5   & {0.66}  & 1 & 2.67 \\ 
  
  Kotlin       & {0.78}   & 3    & {0.77}  & 3 & {0.66}  & 4 & 3.33 \\ 
  
  C       & {0.77}   & 4 & {0.76}  & 4   & {0.66}  & 2 & 3.33 \\ 
  
  Python       & {0.71}   & 6    & {0.79}  & 2  & {0.65}  & 5 & 4.33 \\ 
  
  C++       & {0.76}   & 5    & {0.75}  & 6  & {0.66}  & 3 & 4.67 \\ 
  
  \bottomrule
\end{tabular}
\caption{\label{table:source_ranking_lr} Source languages common for all tasks ranked by mean score across low-resource target languages and three tasks. 
\tagtask is missing because all the target languages for this task are 
high-resource.
Score is the mean of the scores of all target languages for a given source language. 
We see that JavaScript, Java are relatively the best source languages and C++ the worst.}
\end{table*}

The above analysis considers all target languages, including 
high-resource target languages. 
In a more realistic setting, we can ignore high-resource 
target languages because the monolingual option is
available for every high-resource language in every task.
If we repeat the above analysis with only low-resource languages, 
we are left with a smaller set of target languages and only three 
tasks (all target languages for solution domain classification are high-resource).
For the remaining three tasks and six common source languages,
the most transferrable source languages from best to worst are
JavaScript, Java, Kotlin, C, Python, and C++.
JavaScript and Java contribute the most data to CodeT5 pretraining;
Kotlin and C++ are absent~\citep{wang2021codet5}.
For details see Table~\ref{table:source_ranking_lr}.

\begin{table*}[!t]
  \centering
  \begin{tabular}{lccccc}
  \toprule
   & \multicolumn{2}{c}{\clonetask} & \multicolumn{2}{c}{\refinetask} &  \\
  \cmidrule(lr){2-3}\cmidrule(lr){4-5}
   Source & \multicolumn{1}{c}{score}   & rank    & \multicolumn{1}{c}{score} & rank & Mean Rank \\ 
   \cmidrule(lr){1-1}\cmidrule(lr){2-3}\cmidrule(lr){4-5}\cmidrule(lr){6-6}
  
  JavaScript       & {0.81}   & 2    & {0.82}  & 1 & 1.5 \\ 
  
  Scala       & {0.81}   & 1    & {0.77}  & 6 & 3.5 \\ 
  
  Haskell       & {0.81}   & 4    & {0.78}  & 3 & 3.5 \\ 
  
  Kotlin       & {0.78}   & 11 & {0.77}  & 5 & 8 \\ 
  
  D       & {0.81}   & 3    & {0.71}  & 15 & 9 \\ 
  
  Java       & {0.79}   & 9    & {0.75}  & 9 & 9 \\ 
  
  PHP       & {0.79}   & 7    & {0.74}  & 13 & 10 \\ 
  
  Nim       & {0.80}   & 5    & {0.71}  & 16 & 10.5 \\ 
  
  Python       & {0.71}   & 19    & {0.79}  & 2 & 10.5 \\ 
  
  Rust       & {0.79}   & 8    & {0.72}  & 14 & 11 \\ 
  
  Perl       & {0.79}   & 6    & {0.68}  & 17 & 11.5 \\ 
  
  C       & {0.76}   & 15    & {0.76}  & 8 & 11.5 \\ 
  
  Go       & {0.76}   & 16    & {0.77}  & 7 & 11.5 \\ 
  
  C\#       & {0.77}   & 12    & {0.74}  & 12 & 12 \\ 
  
  bf       & {0.55}   & 20    & {0.77}  & 4 & 12 \\ 
  
  Fortran       & {0.78}   & 10    & {0.67}  & 18 & 14 \\ 
  
  Awk       & {0.76}   & 17    & {0.75}  & 11 & 14 \\ 
  
  C++       & {0.76}   & 18    & {0.75}  & 10 & 14 \\ 
  
  Lisp       & {0.77}   & 13    & {0.67}  & 19 & 16 \\ 
  
  Pascal       & {0.77}   & 14    & {0.66}  & 20 & 17 \\ 
  
  \bottomrule
\end{tabular}
\caption{\label{table:source_ranking_clonerefine} Source languages common for \clonetask and \refinetask ranked by mean score across low-resource target languages. 
Score is the mean of the scores of all target languages for a given source language. 
We see that JavaScript, Scala, and Haskell are relatively the best source languages and C++, Lisp, and Pascal the worst.}
\end{table*}

If we consider only \clonetasklower and \refinetasklower, we get a 
much broader set of about 20 common source languages.
Repeating the analysis for these two tasks for low-resource target languages
only, we find once again that JavaScript is the most transferable 
and C++ is among the least.
More details are in Table~\ref{table:source_ranking_clonerefine}.

\begin{tcolorbox}[boxsep=0mm]
  Kotlin is the best source language over all target languages and tasks.
  JavaScript is the best source language for low-resource target languages.
  C++ is a poor source language.
\end{tcolorbox}\vspace*{-3mm}

\paragraph{Language Pair Performance}

\begin{table*}[!t]
  \centering
  \begin{tabular}{l lc lc lc}
      \toprule
       & \multicolumn{2}{c}{\clonetask} & \multicolumn{2}{c}{\refinetask} & \multicolumn{2}{c}{\compiletask} \\ 
      \cmidrule(lr){2-3}\cmidrule(lr){4-5}\cmidrule(lr){6-7}
      Target & \multicolumn{1}{c}{Source}   & score    & \multicolumn{1}{c}{Source}     & score  & \multicolumn{1}{c}{Source} & score \\  
      \cmidrule(lr){1-1}\cmidrule(lr){2-3}\cmidrule(lr){4-5}\cmidrule(lr){6-7}
  
      Crystal       & {Ruby}   & 0.93    & {--}  & --  & {--}  & -- \\ 
  
      Dart       & {Scala}   & 0.93    & {D}  & 0.88   & {--}  & -- \\ 
  
      Swift       & {JavaScript}   & 0.91    & {Kotlin}  & 0.79 & {--}  & -- \\ 
  
      bc       & {Scala}   & 0.91    & {JavaScript}  & 0.87  & {--}  & -- \\ 
  
      TypeScript       & {JavaScript}   & 0.90 & {JavaScript}  & 0.95   & {--}  & -- \\ 
  
      Elixer       & {C}   & 0.89    & {JavaScript}  & 0.83  & {--}  & -- \\ 
  
      Clojure       & {L}   & 0.89    & {bf}  & 0.77  & {--}  & -- \\ 
  
      Visual Basic       & {Scala}   & 0.89    & {JavaScript}  & 0.81  & {--}  & -- \\ 
  
      Julia       & {Ruby}   & 0.89    & {JavaScript}  & 0.81  & {--}  & -- \\ 
  
      Cython       & {Python}   & 0.88    & {Python}  & 0.95  & {--}  & -- \\ 
  
      Lua       & {Ruby}   & 0.88    & {JavaScript}  & 0.87  & {--}  & -- \\ 
  
      Moonscript       & {Scala}   & 0.87    & {JavaScript}  & 0.87  & {--}  & -- \\ 
  
      Scheme       & {Lisp}   & 0.86    & {L}  & 0.83  & {--}  & -- \\ 
  
      COBOL       & {Haskell}   & 0.85    & {Python}  & 0.78  & {--}  & -- \\ 
  
      F\#       & {Haskell}   & 0.84    & {Haskell}  & 0.80  & {--}  & -- \\ 
  
      Sed       & {perl}   & 0.81    & {C}  & 0.89  & {--}  & -- \\ 
  
      Octave       & {ml}   & 0.81    & {Python}  & 0.78  & {--}  & -- \\ 
  
      Bash       & {perl}   & 0.80    & {bf}  & 0.83  & {--}  & -- \\ 
  
      dc       & {perl}   & 0.80    & {--}  & --  & {--}  & -- \\ 
  
      Awk       & {Vim}   & 0.78    & {--}  & --  & {--}  & -- \\ 
  
      C\#       & {--}   & --    & {--}  & --  & {C++}  & 0.68 \\ 
  
      Rust       & {--}   & --    & {--}  & --  & {Python}  & 0.67 \\ 
  
      Ruby       & {--}   & --    & {--}  & --  & {Kotlin, Java}  & 0.67 \\ 
  
      PHP       & {--}   & --   & {--}  & --  & {Kotlin, C++, C}  & 0.67 \\ 
  
      Go       & {--}   & --    & {--}  & --  & {Kotlin}  & 0.67 \\ 
      \bottomrule
  \end{tabular}
  \caption{\label{table:lr_target_best_source} The best
  performing source language for each low resource target language, 
  in each task. For high resource languages, the best performing source language is 
  usually the same language.
  Ruby is high-resource for \clonetask and not a target for \refinetask, 
  hence its results are not show for these two tasks.
  Similarly C\#, Rust, PHP, and Go are high-resource for \clonetask and \refinetask.
  }
  \end{table*}

The best model performance for each task is almost always 
in the monolingual setting.
However, some cross-lingual pairs do stand out and show 
very good model performance, comparable to monolingual.
Another reason to consider individual language pairs is that 
from the perspective of developers working with 
low-resource languages, it would be helpful to know 
which programming languages act as the best 
source language.

For \clonetasklower in \mbox{Figure~\ref{fig:transfer_scores}(a)}, 
a model finetuned on Ruby does almost as well on 
Crystal, a language whose syntax is inspired by Ruby.
\mbox{Figure~\ref{fig:transfer_scores}(a-b)} also shows 
that surprisingly, Haskell does very well on COBOL, much better than 
FORTRAN does on COBOL. 
Lisp does very well 
on Scheme, which can be expected as Scheme is a dialect 
of Lisp.
Similary Python does very well on Cython.
\mbox{Figure~\ref{fig:transfer_scores}(a-d)} also show that 
C-category languages (C, C++, and C\#) do well on each 
other.
Details are in Table~\ref{table:lr_target_best_source}.

\begin{tcolorbox}[boxsep=0mm]
  Some language pairs show particularly good results.
  We need to analyze their features to understand why, motivating
  Section~\ref{sub:Feature_analysis}.
\end{tcolorbox}\vspace*{-3mm}

\paragraph{Dendrograms.}
Dendrograms for each task cluster programming languages
with similar transfer learning performances.
In the \refinetasklower task, syntaxically and semantically related languages are next to each other. For example, C is paired with C++,
JavaScript with TypeScript, Python with Cython, Lisp with Scheme.
On the other hand, relatedness of languages plays a smaller role in
the dendrograms for other tasks.
This might be because while \refinetasklower requires deep language
understanding, other tasks depend more upon other features.
For instance, \citet{ahmed2022multilingual} show that some tasks
depend heavily on identifiers.

\begin{tcolorbox}[boxsep=0mm]
  Relatedness of languages plays a big role for \refinetasklower but
  less so for other tasks.
\end{tcolorbox}\vspace*{-3mm}

\subsection{Performance Prediction}\label{preformance}
Given a task and target language, how should we pick the source language for best performance? Knowing this without having to train models on different source languages can save significant resources. It also provides a basis to study important features that characterize a successful transfer from a language to another.
To predict the performance of the models on the different tasks for different source languages, we train a ranking model based on gradient-boosted decision trees~\citep{ke2017lightgbm}.

We consider the features defined in Table~\ref{table:features},
grouped into four categories.
(1)~\textit{Linguistic features:} general properties
    characterizing a language pair like whether they support the same
    paradigms, the same style of memory management~(garbage collector or not), or the same kind of type-system.
(2)~\textit{Syntactic features:} properties of the syntax of a
    language pair measured by overlap in the counts of certain token
    types, as determined using the Pygments library on the corpus of
    code in the given languages.
(3)~\textit{Dataset-specific features:} properties of the problems
    associated with code samples for the language pair in the online
    coding judge website from which the dataset came.
(4)~\textit{Model-specific features:} whether the model saw the
    source, the target, or both languages during pretraining.
The linguistic features of each language are given Table~\ref{tab:linguistic-features}.

\begin{table}[t]
\centerline{\begin{tabular}{lcl}
\toprule
  Feature & Measure & Description\\\hline
\multicolumn{3}{l}{Linguistic Features}\\\hline
  Object oriented     &  e  & Is the language object oriented?\\
  Type strength       &  e  & Is the language strongly or weakly typed?\\
  Type checking       &  e  & Is the language typed statically or dynamically?\\
  Type safety         &  e  & Is the language type safe?\\
  Garbage collection  &  e  & Does the language use garbage collector?\\
  Standardized        &  e  & Is the language standardized by a committee?\\
  Expression of types &  e  & Are the types written explicitly?\\
  Paradigm            &  o  & Paradigms supported (e.g., functional, imperative)\\
  Type compatibility  &  o  & Is language nominally-typed or structurally-typed?\\
  Parameter passing   &  o  & How are params.\ passed (e.g.\ by value, by name)?\\\hline
\multicolumn{3}{l}{Syntactic Features}\\\hline
  Name                &  o  & Number of names which overlap\\
  Text                &  o  & Number of text data which overlap\\
  Keyword             &  o  & Number of keywords which overlap\\
  Literal             &  o  & Number of literals which overlap\\
  Punctuation         &  o  & Number of punctuation signs which overlap\\
  Operator            &  o  & Number of operators which overlap\\
  Comment             &  o  & Number of comment tokens which overlap\\
  Syntax              &  o  & Number of AST nodes which overlap\\
  Tokens              &  o  & Number of tokens which overlap\\\hline
\multicolumn{3}{l}{Dataset-Specific Features}\\\hline
  Difficulty          & $x$ & Average difficulty of dataset problems\\
  Length              & $x$ & Average number of tokens\\
  Time limit          & $x$ & Average time limit of dataset problems\\
  Memory limit        & $x$ & Average memory limit of dataset problems\\\hline
\multicolumn{3}{l}{Model-Specific Features}\\\hline
  Pretrained          &  s  & Source language is included during pretraining\\
  Pretrained          &  t  & Target language is included during pretraining\\
  Pretrained          &  b  & Both source and target languages in pretraining\\
\bottomrule
\end{tabular}}
\caption{\label{table:features}Features of language pairs. Column ``Measure'' indicates how the feature is computed: (e)quivalence, (o)verlap, (s)ource, (t)arget, (b)oth source and target, (rd) for relative difference. For dataset specific features, ($x$) can be s, t, or rd. Details of the linguistic features are in Table~\ref{tab:linguistic-features}.}
\end{table}


    \afterpage{\begin{landscape}
    \begin{table}[h]
        \resizebox{9.5in}{!}
        {\begin{tabular}{l|lccccccccc}
        Language & Paradigm                                                                                               & \begin{tabular}[c]{@{}c@{}}Object\\Oriented\end{tabular}  & Standardized & \begin{tabular}[c]{@{}c@{}}Type\\Strength\end{tabular}  & \begin{tabular}[c]{@{}c@{}}Type\\Safety\end{tabular}  &  \begin{tabular}[c]{@{}c@{}}Expression\\of Types\end{tabular}   &  \begin{tabular}[c]{@{}c@{}}Type\\Compatibility\end{tabular}   & \begin{tabular}[c]{@{}c@{}}Type\\Checking\end{tabular} & Parameter Passing                                                    & \begin{tabular}[c]{@{}l@{}}Garbage\\Collection\end{tabular}\\ \hline
 
Awk & \docircle{blue}{green}{white}{S}\docircle{blue}{black}{white}{D}\docircle{orange}{pink}{black}{P} & \xmark & \cmark & strong & \xmark & implicit &  & dynamic & by value,   by reference & \cmark \\
Bf & \docircle{green}{blue}{black}{I}\docircle{yellow}{red}{black}{S}\docircle{black}{blue}{white}{E} & \xmark & \xmark & weak & \xmark & implicit &  & dynamic &  & \xmark \\
C & \docircle{green}{blue}{black}{I}\docircle{yellow}{red}{black}{S} & \xmark & \cmark & weak & \xmark & explicit & nominal & static & by value,  by reference & \xmark \\
Clojure & \docircle{blue}{purple}{white}{C}\docircle{orange}{green}{black}{A}\docircle{orange}{yellow}{black}{F}\docircle{orange}{purple}{black}{L}\docircle{red}{purple}{black}{P}\docircle{red}{orange}{black}{M} & \xmark & \xmark & strong & \cmark & implicit & structural & dynamic & by value & \cmark \\
COBOL & \docircle{green}{blue}{black}{I}\docircle{orange}{blue}{black}{G}\docircle{blue}{red}{white}{O}\docircle{orange}{pink}{black}{P} & \cmark & \cmark & strong & \cmark & explicit & nominal & static & by value,  by reference & \xmark \\
C++ & \docircle{green}{blue}{black}{I}\docircle{orange}{blue}{black}{G}\docircle{orange}{yellow}{black}{F}\docircle{blue}{red}{white}{O}\docircle{orange}{pink}{black}{P}\docircle{orange}{red}{black}{M} & \cmark & \cmark & strong & \cmark & explicit & nominal,  structural & static & by value,  by reference & \xmark \\
C\# & \docircle{green}{blue}{black}{I}\docircle{yellow}{red}{black}{S}\docircle{blue}{purple}{white}{C}\docircle{yellow}{brown}{black}{R}\docircle{orange}{blue}{black}{G}\docircle{orange}{yellow}{black}{F}\docircle{blue}{red}{white}{O}\docircle{blue}{black}{white}{E}\docircle{yellow}{red}{black}{T} & \cmark & \cmark & strong & \cmark & implicit & nominal & static & by value,  by reference & \cmark \\
D & \docircle{green}{blue}{black}{I}\docircle{orange}{blue}{black}{G}\docircle{blue}{red}{white}{O} & \cmark & \xmark & strong & \cmark & explicit & nominal,  structural & static & by value,  by reference & \cmark \\
Dart & \docircle{green}{blue}{black}{I}\docircle{yellow}{brown}{black}{R}\docircle{orange}{yellow}{black}{F}\docircle{blue}{red}{white}{O} & \cmark & \xmark & strong & \xmark & inferred & nominal & static & by value & \cmark \\
Elixir & \docircle{blue}{purple}{white}{C}\docircle{orange}{yellow}{black}{F}\docircle{blue}{brown}{white}{D}\docircle{yellow}{red}{black}{P} & \xmark & \xmark & strong & \xmark & implicit & structural & dynamic & by value & \cmark \\
Fortran & \docircle{green}{blue}{black}{I}\docircle{yellow}{red}{black}{S}\docircle{yellow}{orange}{black}{A}\docircle{orange}{blue}{black}{G}\docircle{blue}{red}{white}{O}\docircle{orange}{pink}{black}{P} & \cmark & \cmark & strong & \cmark & explicit & nominal & static & by reference & \xmark \\
F\# & \docircle{green}{blue}{black}{I}\docircle{blue}{purple}{white}{C}\docircle{yellow}{brown}{black}{R}\docircle{orange}{green}{black}{A}\docircle{orange}{yellow}{black}{F}\docircle{blue}{red}{white}{O}\docircle{blue}{blue}{white}{M} & \cmark & \xmark & strong & \cmark & implicit & nominal & static & by value & \cmark \\
Go & \docircle{green}{blue}{black}{I}\docircle{blue}{purple}{white}{C}\docircle{blue}{red}{white}{O} & \cmark & \xmark & strong & \cmark & explicit & nominal & static & by value & \cmark \\
Haskell & \docircle{orange}{yellow}{black}{F} & \xmark & \cmark & strong & \cmark & inferred & structural & static & by name & \cmark \\
Java & \docircle{green}{blue}{black}{I}\docircle{blue}{purple}{white}{C}\docircle{yellow}{brown}{black}{R}\docircle{orange}{blue}{black}{G}\docircle{orange}{yellow}{black}{F}\docircle{blue}{red}{white}{O} & \cmark & \cmark & strong & \cmark & explicit & nominal & static & by value & \cmark \\
Julia & \docircle{yellow}{brown}{black}{R}\docircle{orange}{yellow}{black}{F}\docircle{blue}{red}{white}{O}\docircle{orange}{pink}{black}{P}\docircle{yellow}{red}{black}{M} & \cmark & \xmark & strong & \cmark & inferred & nominal & dynamic & by value,   by reference & \cmark \\
JavaScript & \docircle{green}{blue}{black}{I}\docircle{orange}{yellow}{black}{F}\docircle{blue}{red}{white}{O}\docircle{orange}{pink}{black}{P}\docircle{blue}{black}{white}{E} & \cmark & \cmark & weak & \xmark & implicit & structural & dynamic & by value & \cmark \\
Kotlin & \docircle{green}{blue}{black}{I}\docircle{blue}{purple}{white}{C}\docircle{yellow}{brown}{black}{R}\docircle{orange}{blue}{black}{G}\docircle{orange}{yellow}{black}{F}\docircle{blue}{red}{white}{O}\docircle{yellow}{orange}{black}{B}\docircle{yellow}{red}{black}{D} & \cmark & \xmark & strong & \cmark & explicit & nominal & static & by value & \cmark \\
Lisp & \docircle{yellow}{brown}{black}{R}\docircle{orange}{blue}{black}{G}\docircle{orange}{yellow}{black}{F}\docircle{blue}{red}{white}{O}\docircle{orange}{pink}{black}{P} & \cmark & \cmark & strong & \cmark & implicit & nominal,  structural & dynamic & by value & \cmark \\
Lisp & \docircle{yellow}{brown}{black}{R}\docircle{orange}{yellow}{black}{F}\docircle{blue}{red}{white}{O}\docircle{orange}{pink}{black}{P} & \cmark & \xmark & strong & \cmark & implicit & nominal,  structural & dynamic & by value & \cmark \\
Lua & \docircle{blue}{green}{white}{S}\docircle{green}{blue}{black}{I}\docircle{yellow}{brown}{black}{R}\docircle{orange}{yellow}{black}{F}\docircle{blue}{red}{white}{O}\docircle{orange}{pink}{black}{P} & \xmark & \xmark & strong & \cmark & implicit & structural & dynamic & by value & \cmark \\
ocaml & \docircle{green}{blue}{black}{I}\docircle{orange}{yellow}{black}{F}\docircle{blue}{red}{white}{O}\docircle{orange}{red}{black}{M} & \cmark & \xmark & strong & \cmark & inferred & structural & static & by value & \cmark \\
Moonscript & \docircle{blue}{red}{white}{O} & \cmark & \xmark & weak & \xmark & implicit & structural & dynamic & by value & \cmark \\
Nim & \docircle{green}{blue}{black}{I}\docircle{blue}{purple}{white}{C}\docircle{orange}{yellow}{black}{F}\docircle{blue}{red}{white}{O}\docircle{orange}{pink}{black}{P}\docircle{blue}{orange}{white}{C} & \cmark & \xmark & strong & \cmark & explicit & nominal,  structural & static & by value,  by reference & \cmark \\
Octave & \docircle{blue}{red}{white}{O} & \cmark & \xmark & weak & \xmark & implicit &  & dynamic & by value & \cmark \\
Pascal & \docircle{green}{blue}{black}{I}\docircle{orange}{pink}{black}{P} & \xmark & \cmark & strong & \cmark & explicit & nominal & static & by reference,  by value & \xmark \\
PHP & \docircle{green}{blue}{black}{I}\docircle{yellow}{brown}{black}{R}\docircle{orange}{yellow}{black}{F}\docircle{blue}{red}{white}{O}\docircle{orange}{pink}{black}{P} & \cmark & \xmark & weak & \xmark & implicit & nominal & dynamic & by value,  by reference & \cmark \\
Python & \docircle{yellow}{red}{black}{S}\docircle{yellow}{brown}{black}{R}\docircle{orange}{yellow}{black}{F}\docircle{blue}{red}{white}{O}\docircle{orange}{pink}{black}{P} & \cmark & \xmark & strong & \cmark & implicit & structural & dynamic & by value & \cmark \\
Cython & \docircle{green}{blue}{black}{I}\docircle{yellow}{brown}{black}{R}\docircle{orange}{blue}{black}{G}\docircle{orange}{yellow}{black}{F}\docircle{blue}{red}{white}{O} & \cmark & \xmark & strong & \xmark & implicit & nominal,  structural & dynamic & by value & \cmark \\
Ruby & \docircle{green}{blue}{black}{I}\docircle{yellow}{brown}{black}{R}\docircle{orange}{yellow}{black}{F}\docircle{blue}{red}{white}{O} & \cmark & \cmark & strong & \xmark & implicit & structural & dynamic & by value & \cmark \\
Rust & \docircle{green}{blue}{black}{I}\docircle{yellow}{red}{black}{S}\docircle{blue}{purple}{white}{C}\docircle{orange}{blue}{black}{G}\docircle{orange}{yellow}{black}{F}\docircle{blue}{red}{white}{O} & \cmark & \xmark & strong & \cmark & explicit & nominal & static & by value,  by reference & \xmark \\
Scala & \docircle{green}{blue}{black}{I}\docircle{blue}{purple}{white}{C}\docircle{orange}{yellow}{black}{F}\docircle{blue}{red}{white}{O} & \cmark & \cmark & strong & \cmark & implicit & nominal,  structural & static & by value,  by name & \cmark \\
Scheme & \docircle{green}{blue}{black}{I}\docircle{orange}{yellow}{black}{F}\docircle{blue}{red}{white}{O} & \cmark & \cmark & strong & \cmark & implicit & structural & dynamic & by value & \cmark \\
Bash &  & \xmark & \cmark & weak & \xmark & implicit &  & dynamic & by value,   by reference & \cmark \\
Swift & \docircle{green}{blue}{black}{I}\docircle{blue}{purple}{white}{C}\docircle{orange}{yellow}{black}{F}\docircle{blue}{red}{white}{O}\docircle{yellow}{orange}{black}{B}\docircle{yellow}{red}{black}{D} & \cmark & \xmark & strong & \cmark & inferred & nominal & static & by value & \cmark \\
Typescript & \docircle{green}{blue}{black}{I}\docircle{orange}{blue}{black}{G}\docircle{orange}{yellow}{black}{F}\docircle{blue}{red}{white}{O} & \cmark & \xmark & strong & \xmark & explicit & structural & static &  by value & \cmark \\
Visual Basic & \docircle{green}{blue}{black}{I}\docircle{yellow}{red}{black}{S}\docircle{yellow}{brown}{black}{R}\docircle{orange}{blue}{black}{G}\docircle{blue}{red}{white}{O}\docircle{blue}{black}{white}{E}\docircle{yellow}{red}{black}{D} & \cmark & \xmark & strong & \cmark & implicit & nominal & static & by reference,  by value & \cmark \\
Vim & \docircle{green}{blue}{black}{I}\docircle{orange}{yellow}{black}{F}\docircle{blue}{red}{white}{O} & \cmark & \xmark & weak & \xmark & implicit & structural & dynamic & by value & \cmark \\
\end{tabular}}
\\
~\hfill
\begin{minipage}{9.5in}\vspace{5mm}\caption{\label{tab:linguistic-features}Linguistic features. Paradigms: 
\docircle{blue}{green}{white}{S}: scripting
\docircle{green}{blue}{black}{I}: imperative
\docircle{yellow}{red}{black}{S}: structured
\docircle{blue}{pink}{black}{A}: aspect-oriented
\docircle{blue}{purple}{white}{C}: concurrent
\docircle{yellow}{orange}{black}{A}: array
\docircle{blue}{black}{white}{D}: data-driven
\docircle{black}{blue}{white}{E}: esoteric
\docircle{yellow}{brown}{black}{R}: reflective
\docircle{orange}{blue}{black}{G}: generic
\docircle{orange}{green}{black}{A}: agent-oriented
\docircle{orange}{yellow}{black}{F}: functional
\docircle{blue}{red}{white}{O}: object-oriented
\docircle{orange}{pink}{black}{P}: procedural
\docircle{orange}{purple}{black}{L}: logic
\docircle{blue}{orange}{white}{C}: compiled
\docircle{blue}{black}{white}{E}: event-driven
\docircle{blue}{blue}{white}{M}: metaprogramming
\docircle{blue}{brown}{white}{D}: distributed
\docircle{orange}{red}{black}{M}: modular
\docircle{yellow}{orange}{black}{B}: block structured
\docircle{red}{purple}{black}{P}: pipeline
\docircle{red}{orange}{black}{M}: macro
\docircle{yellow}{red}{black}{M}: multistaged
\docircle{yellow}{red}{black}{P}: process-oriented
\docircle{yellow}{red}{black}{T}: task-driven
\docircle{yellow}{red}{black}{D}: declarative
\docircle{orange}{yellow}{black}{P}: process-driven
}\end{minipage}
\end{table}
\end{landscape}}

To predict the top source languages for a given task and target language $L_t$ in the set of target languages $T$, we train a ranker using the LightGBM \citep{ke2017lightgbm} implementation of the LambdaRank algorithm~\citep{burges_2010}. The model takes  features in Table~\ref{table:features} as inputs and scores source languages $L_s$ in terms of their relevance to the target $L_t$.  Our ranker uses a boosting ensemble of 100 decision trees with 16 leaves each. We consider the normalized discounted cumulative gain score (NDCG@K) at $K=3$ as our evaluation metric. We evaluate the model using leave-one-out (LOO) cross-validation on the set $T$. For each target language $L_t$, we train a ranker to predict rankings of different sources for each language in $T$ leaving out the source ranking for $L_t$ as a test set. For each fold, we compute the NDCG@3 score on the test set.

\begin{figure}[t]
\centering
\centerline{\includegraphics[width=0.35\columnwidth,]{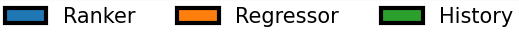}}
\centerline{\includegraphics[width=.6\columnwidth,]{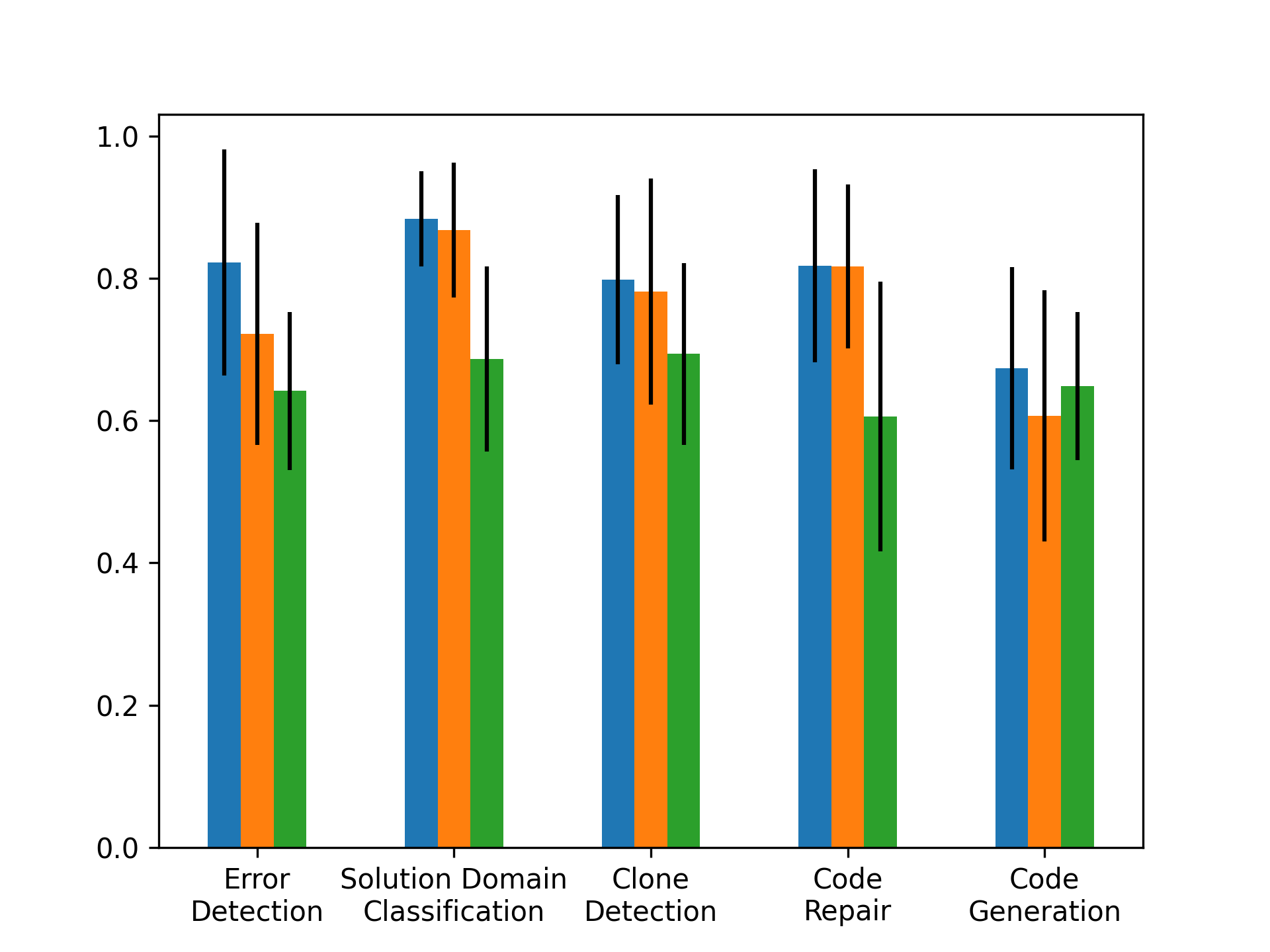}}

\caption{NDCG@3 scores for different rankers and tasks corresponding to LOO evaluation over the set of target languages\label{fig:rankers}. {The first four tasks are evaluated on ranking predictions on CodeT5 transfer experiments, while code generation is evaluated over ranking predictions on \LlamaInstr.}}
\end{figure}

We compare the performance of our ranker with a regression-based ranker and a history ranker.
The regression-based ranker uses a LightGBM regressor with the same hyperparameter settings as our ranker and selects the top $K$ source languages with the highest predicted scores. 
The history ranker selects the top $K$ source languages with the shortest path to the given target language in a programming languages history graph\footnote{\url{http://rigaux.org/language-study/diagram.html}} (contracted to merge different versions of the same language).
Figure~\ref{fig:rankers} shows the mean and standard deviation of NDCG@3 using LOO on~$T$.
While the error bars are overlapping, 
both regression and rankers outperform the history ranker on all tasks, indicating the importance of considering a variety of features for predicting top-performing sources rather than simplified heuristics. 
Our model outperforms baseline rankers on all tasks with the exception of \clonetasklower, for which it shows slightly lower performance than the regression model. This indicates that ranking is the best suited method for predicting top performing sources. Based on these results, we use our ranker for the feature analysis in RQ3.

\begin{tcolorbox}[boxsep=0mm]
  Rankers outperform regressors for predicting top-performing sources.
  Using historical relationships among programming languages alone is insufficient.
\end{tcolorbox}\vspace*{-3mm}

\subsection{Feature analysis}\label{sub:Feature_analysis}

\begin{figure}
\centerline{\includegraphics[width=0.95\textwidth]{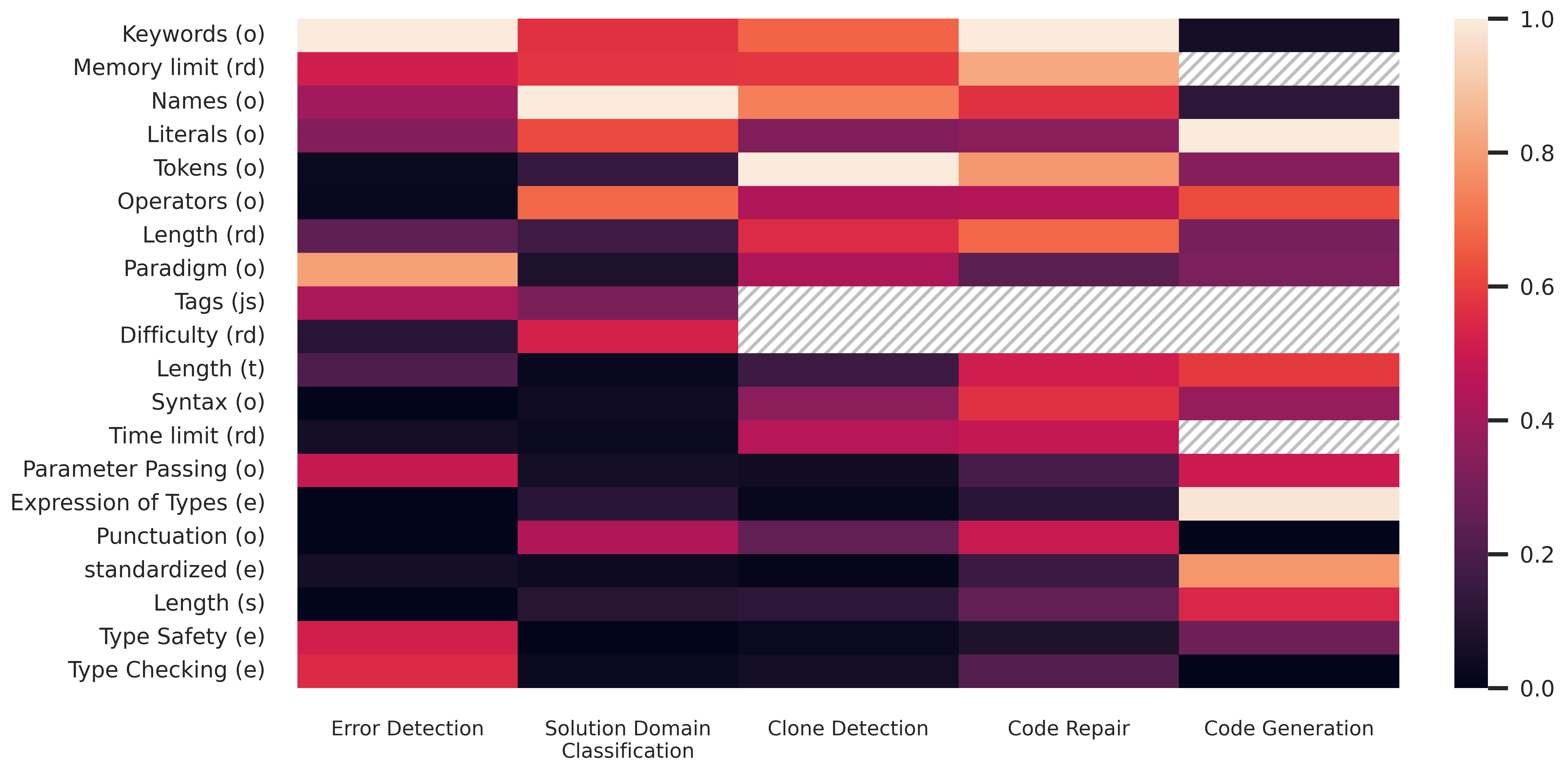}}
\caption{\label{fig:shap_pw}Normalized SHAP values aggregated by tasks for the features defined in Table~\ref{table:features}. The features are sorted by mean rank. Grayed-out features were unavailable for the given task. {The first four tasks are evaluated on ranking predictions on CodeT5 transfer experiments, while code generation is evaluated over ranking predictions on \LlamaInstr.}}
\end{figure}

While Section~\ref{transfer_analysis} explored how well transfer learning
performs for different languages on different tasks, we would like to
dig deeper into \emph{why} they perform that way.
That is, how do the features of language pairs affect transfer?
Which characteristics of the source and target language best explain
the variation of transfer across tasks, the dependence of transfer on
source and target pairing, as well as the superior transfer supported by
specific sources?
These questions can be answered by measuring feature importance.

The state-of-the-art technique for measuring feature importance, SHAP values~\citep{lundberg_lee_2017}, considers all possible subsets $S\subseteq F$ of features. Then, the importance of a given feature is the difference it makes to model performance in combination with all possible sets of other features. In the context of ranking different source languages for cross-lingual transfer, SHAP values measure how different features contribute to transfer. Figure~\ref{fig:shap_pw} shows SHAP values computed based on the ranking models evaluated in Section~\ref{preformance}. For presentation purposes, it rescales the values so the most important feature for each task has importance~1.

\paragraph{Task-dependent feature importance.}
While previous works on transfer learning across programming languages
emphasize the importance of specific features for
cross-lingual transfer, the lack of comparison on different tasks
limits their conclusions.
The supported cross-task analysis demonstrates a task-dependent
feature importance whereby different features contribute differently
to cross-lingual transfer depending on the task.
For example, the top features for the \tagtasklower task are
\textit{Difficulty}, \textit{Literals}, \textit{Names}, and
\textit{Operators}.
Being able to predict the solution domain for a code solution requires
some knowledge of the underlying problem for which the tags are an
attribute.
The difficulty score is another attribute of the problem.
Problems with similar difficulty scores require similar algorithms.
The literals, names, and operators are other indicators of the
algorithms used in a code sample.
In comparison, the most significant features for the \clonetasklower task
are \textit{Token}, \textit{Names}, and \textit{Keywords}.
Detecting a clone requires different skills than classifying domains.
A deeper understanding of the code semantics irrespective of the
similar algorithms used is needed.
The overlaps in names, keywords, and more generally tokens is key for
understanding the semantics of code~\citep{ahmed2022multilingual}.

\begin{tcolorbox}[boxsep=0mm]
  Prior work on transfer learning in natural languages~\citep{lin_et_al_2019}
  showed feature importances to vary greatly across tasks.
  We confirm this for programming languages.
\end{tcolorbox}\vspace*{-3mm}
\paragraph{Range of important features.}
Different tasks seem to not only focus on different features, but also
on a different number of features.
While the feature importance for \tagtasklower and \compiletasklower is
concentrated on a small number of features, it is
more spread out for \clonetasklower and \refinetasklower.
For example, the \refinetasklower task requires a transfer of knowledge
from overlapping keywords, names, and more generally different tokens
from a source language for fixing a code in a target language.
On the other hand, \tagtasklower focuses on fewer features, related to the
problems attributes.

\begin{tcolorbox}[boxsep=0mm]
  Higher-level tasks seem to draw upon more features to
  best understand the code.
\end{tcolorbox}\vspace*{-3mm}

\paragraph{Comparison across categories.}
Comparing the importance of different feature categories across tasks,
we observe a relatively higher significance of syntactic features as
demonstrated by the top two features \textit{Keywords} and
\textit{Names}.
Compared to transfer learning across natural
languages, model-specific and linguistics features
seem to carry less significance for transfer in programming languages.
While \citet{de2022make} find the potential of cross-lingual
transfer dropping significantly for sources and targets that are not
seen by the model, our finding brings hope to not only transferring
learning to low-resource target languages, but even finetuning the
model on unseen languages that could serve as good sources, such as Kotlin.

\begin{tcolorbox}[boxsep=0mm]
  Overall, the most important features across the four tasks
  are keywords and names.
\end{tcolorbox}\vspace*{-3mm}

\paragraph{Within-category differences.}
While different categories seem to have a higher impact on the model
predictions, the impact of single features within a category varies
across task.
Our results are consistent with prior work that found \textit{Names}
and \textit{Keywords} to be the most important features for
cross-lingual transfer in certain tasks~\citep{ahmed2022multilingual}.
However, half of the tasks show keywords to be more important than
names, while the other half has the opposite order of importance.

\subsection{{Few-shot prompting}}\label{sub:scaling}

{We examine cross‑lingual transfer learning for programming languages by pairing low‑ and high‑resource languages across classification and code‑repair tasks. Whereas our initial study employed CodeT5‑base, we now replicate a representative subset of experiments with the larger \LlamaInstr model to test the generalizability of our conclusions to contemporary large‑scale LLMs. For evaluation we adopt the few‑shot prompting scheme described in Section~\ref{sec: approach}; this technique has been shown to boost out‑of‑domain code‑generation quality for previously unseen languages and to markedly reduce the compilation and parsing failures that arise when the model lacks explicit exposure to a target language \citep{athiwaratkun2023multilingual}.}

{We investigate whether rankers constructed from the features introduced earlier can accurately predict the best source language(s) for few‑shot prompting. We compare different rankers explored in the previous section and test whether our ranker model can generalize to larger models. We train a ranker model on the same features described in Table \ref{table:features} on Pass@1 scores. Figure~\ref{fig:rankers} shows mean NDCG@3 scores. The learned ranker still surpasses other rankers and heuristic baselines, underscoring that considering multiple features is essential for accurately identifying the top source languages. SHAP analysis further reveals that expression of types and overlap in literals dominate the other features in predicting top performing sources. Languages whose type systems are richly expressed in code provide stronger semantic signals that transfer directly for code generation of other similarly typed languages solutions. Meanwhile a high lexical overlap in literals such as numeric and string constants offers immediate token‑level anchors that reduce out‑of‑vocabulary risk during generation. Together, these two features jointly encode semantic structure and surface‑level similarity, making them the most dependable indicators of best source languages for code generation.}

\section{Limitations}
\label{sec: limitations}

This section describes the limitations of our study, including the rationale for key design choices and the steps we took to address those limitations.

\subsection{External validity}

One limitation is that, to keep the large number of experiments feasible and reduce their carbon footprint, we constrained our fine-tuning experiments to a moderately sized model. We address this by using CodeT5 \citep{wang2021codet5}, 
a commonly adopted model in AI-for-Code research. Even if a larger model were used, insights from one model would not necessarily generalize to another. 
Accordingly, we included model-specific features in our study to account for potential model dependence. 
These features are described in Table~\ref{table:features} and analyzed as part of Section~\ref{sub:Feature_analysis}.
{Despite model-specific features, we do acknowledge that experimenting with different models, not just of varying size but also architecture, would have made our study more robust.
Because the source and target languages are usually different and come in many combinations, we do not necessarily run the risk of overfitting.
However, using models of varying strength could have exposed a broader range of more interesting patterns which would have made our study richer and more general.
To mitigate this limitation, we conduct separate experiments with a larger model and few-shot prompting.}

Another limitation is that, to cover multiple languages, we limited ourselves to only five tasks. 
We address this by selecting tasks that span a broad range of difficulty and demonstrate diversity, 
as indicated by their differing performance and feature importance. 
Nonetheless, our study includes more tasks than any previous cross-lingual transfer study on programming languages.

A further limitation is that our datasets include synthetic elements—for instance, 
\mbox{type-IV} clones \citep{roy_cordy_2007} in the \clonetasklower data or fault injection in the \refinetasklower data. 
However, using exclusively natural data imposes a demanding constraint that few academic studies meet, and it is also uncommon in industrial practice.
To address this, we explicitly measure the importance of dataset-specific features for each language pair.

Another limitation is that we focused on the setting of zero data in the target language.
This represents the lowest barrier to entry, as gathering even a small number of examples can be tricky.
Therefore, we defer exploring scenarios that include some target-language data to future work.

\subsection{Internal validity}

We believe that our study avoids one common internal
validity limitation, which revolves around insufficient runs. Across all experiments, each source-target pair is evaluated only once. But thanks to the sheer number of language pairs for each task, we get thousands of different experiments, whose results are consistent with each other. Moreover, each of these experiments involves fine-tuning with tens of thousands of samples, giving us confidence that the choice of samples is not a limitation either. However, having results based on single runs does limit our ability to asses variability in experiments. Since our study is the first of its kind at this scale, it lacks directly comparable baselines. We mitigate this limitation by noting that, wherever comparisons are possible, our results corroborate those of narrower studies—for instance, prior work on transfer for identifier-centric tasks~\citep{ahmed2022multilingual}.

\section{Conclusion}
\label{sec: conclusion}

We perform a systematic and extensive study of LLM transfer learning covering up to 41 programming languages, across 5 tasks including both classification and generation. Programming languages covered include several low-resource but often still widely-used languages. Cross-lingual transfer learning performs much better than zero-shot with an LLM that has been pre-trained on code. One interesting finding is that even a language not seen during pre-training, like Kotlin, can be good fine-tuning source language across several target languages and tasks. On the other hand, certain languages that are often used extensively to pre-train LLMs, like C++ and Python, are worse source or target languages relative to others.

To understand relative differences in cross-lingual performance between different programming languages, we define several linguistic, syntactic, dataset, and model-specific features of language pairs and then analyze the feature importance for a model that predicts transfer performance. We show how different features are needed for ranking source languages, compared to ad-hoc heuristics such as overlap on names. To explain how a language that was unseen during pre-training, like Kotlin, can be a good source language, we show that seen language features are less significant compared to dataset, linguistic, and syntactic source language features. Similar to previous work of~\cite{ahmed2022multilingual}, we show that keywords and names are top features on average for CodeT5. On the other hand, unlike previous work, we cover more languages and more diverse tasks, and find that feature importance vary strongly across tasks. Replicating a representative subset of experiments with a larger model under few‑shot prompting confirms that these feature hierarchies and the performance prediction gains they enable generalize to state‑of‑the‑art LLMs. Overall, we believe that this paper sheds light on how learning transfers among programming languages, and that this ultimately leads to better models to assist users of those languages, particularly low-resource ones.

\section{Data Availability}
\label{sec: data_availability}
The experiments are based on the publicly available CodeT5-base (220M parameters) model~\citep{wang2021codet5} and the open-sourced datasets CodeNet~\citep{puri2021codenet} and XCodeEval~\citep{khan_et_al_2023}. Our code is available at \url{https://github.com/baltaci-r/XPL}.

\bibliography{custom}
\bibliographystyle{tmlr}

\end{document}